%% file: iclr2027_conference.tex
\documentclass[10pt,a4paper,logo]{paper}
\usepackage{fix-cm}
\usepackage[all]{hypcap}
\usepackage[authoryear,round]{natbib}
\usepackage{hyperref}
\usepackage{url}
\usepackage{fancyvrb}
\usepackage{booktabs}
\usepackage{fvextra}
\usepackage[capitalise]{cleveref}
\usepackage{subcaption}
\usepackage{multirow}
\usepackage{makecell}
\usepackage{wrapfig}
\usepackage{algorithm}
\usepackage{algpseudocode}
\usepackage{arydshln}
\usepackage{tabularx}
\usepackage[most]{tcolorbox}
\usepackage{mdframed}

\input{math_commands.tex}

\newcommand{\modelname}[1]{\textsc{FlyBy}}
\newcommand{\dgr}[1]{\textcolor{gray!70!black}{#1}}

\definecolor{Red}{rgb}{0.768, 0.054, 0.054}
\definecolor{citeblue}{RGB}{0,114,178}
\definecolor{title_blue}{HTML}{204899}
\definecolor{black50}{gray}{0.5}
\definecolor{celestialblue}{rgb}{0.29, 0.59, 0.82}
\definecolor{ceruleanblue}{rgb}{0.16, 0.32, 0.75}
\definecolor{barblue}{rgb}{0.1412, 0.3647, 0.5686}

\hypersetup{
    colorlinks=true,
    citecolor=barblue,
    linkcolor=Red,
    urlcolor=ceruleanblue,
}

\crefname{table}{Tab.}{Tabs.}
\Crefname{table}{Tab.}{Tabs.}

\newmdenv[
    linecolor=black,
    linewidth=0.5pt,
    backgroundcolor=yellow!10!white,
    roundcorner=4pt
]{promptbox}

\definecolor{takeawayblue}{RGB}{51,92,141}
\definecolor{takeawaybg}{RGB}{248,250,252}

\newtcolorbox{takeaway}{
    enhanced,
    colback=takeawaybg,
    colframe=takeawayblue,
    boxrule=0.5pt,
    arc=0.5mm,
    left=6pt,
    right=6pt,
    top=4pt,
    bottom=4pt,
    before skip=8pt,
    after skip=8pt,
}

\title{
Knowing When Thinking Is Not Enough: \\
Teaching Small Reasoning Models to Reason \\
Beyond Their Parametric Knowledge
}
\newcommand{\shorttitle}{Knowing When Thinking Is Not Enough}

\newcommand{\authoremails}[1]{%
  {\normalfont\fontsize{9}{11}\selectfont
   \makebox[1.3em][c]{\raisebox{-0.15em}{\includegraphics[height=1em]{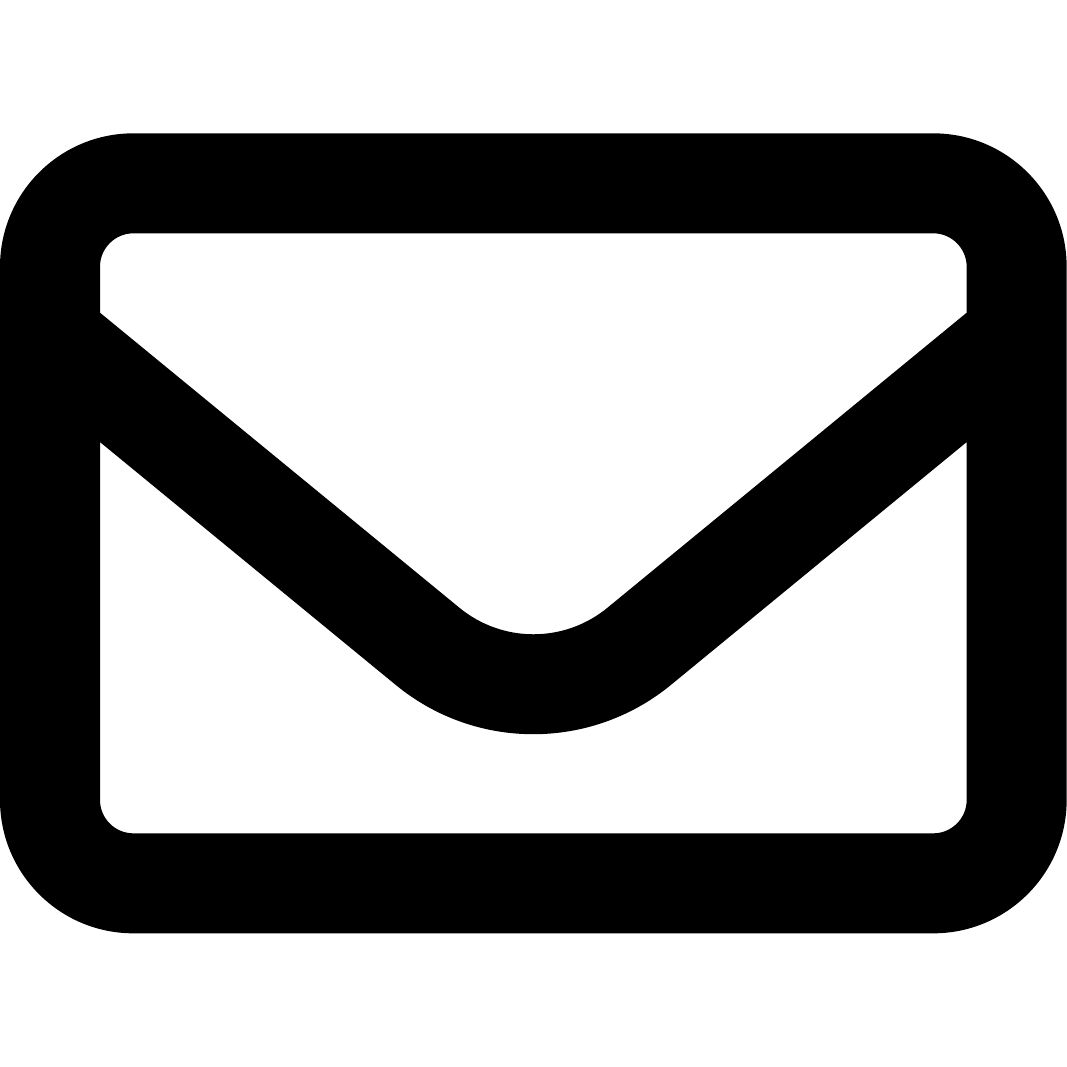}}}~\texttt{#1}}}
\newcommand{\authorcode}[1]{%
  {\normalfont\fontsize{9}{11}\selectfont
   \makebox[1.3em][c]{\raisebox{-0.15em}{\includegraphics[height=1em]{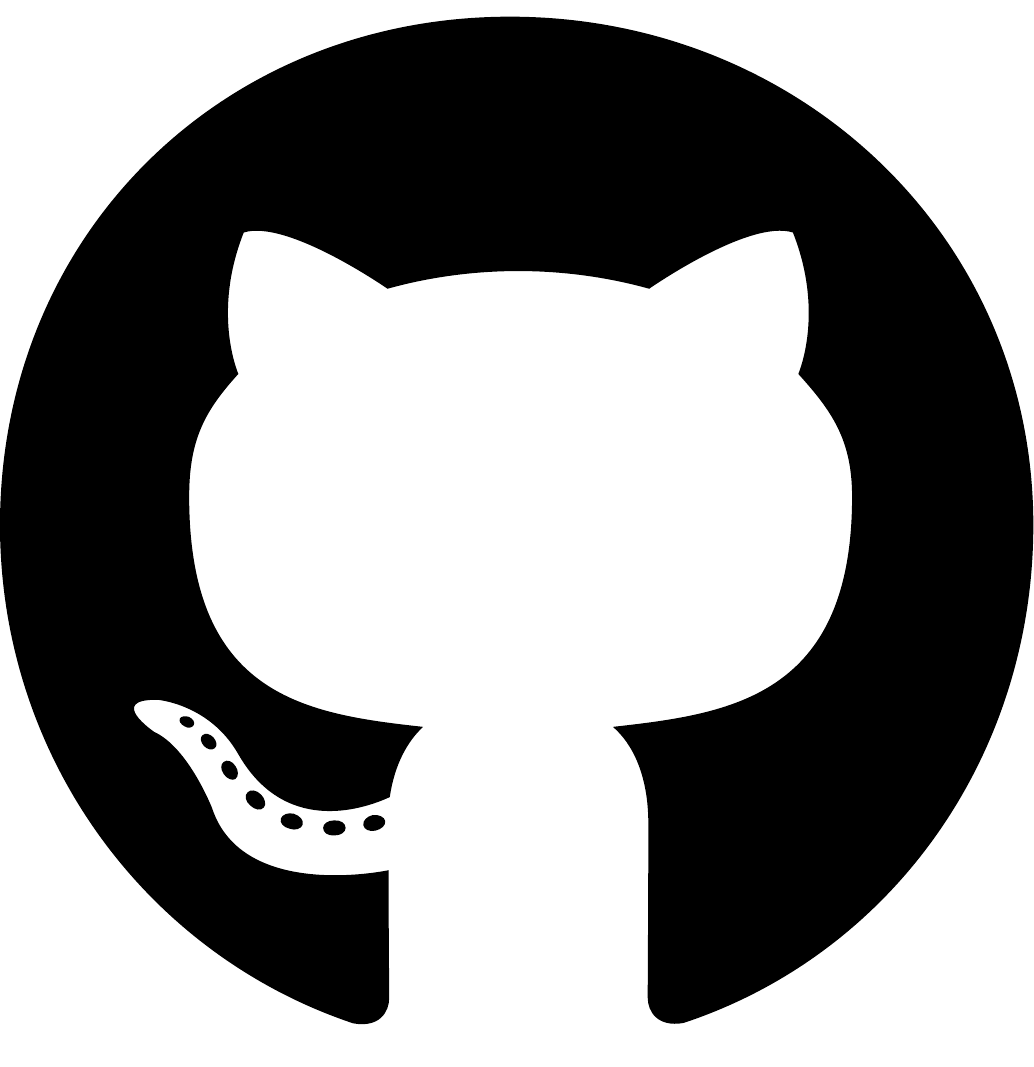}}}~\url{#1}}}

\author{
Chanuk Lee\textsuperscript{1},
Minki Kang\textsuperscript{1},
Sangwoo Park\textsuperscript{1},
Woongyeong Yeo\textsuperscript{1},
Jinheon Baek\textsuperscript{1,$\dagger$},
Sung Ju Hwang\textsuperscript{1,2,$\dagger$} \\
\textsuperscript{1}KAIST \quad
\textsuperscript{2}DeepAuto.ai \quad
($\dagger$: Equal advising)
\\
\authoremails{\{tallyforce, jinheon.baek, sungju.hwang\}@kaist.ac.kr} \\
\authorcode{https://github.com/tally0818/FlyBy}
}

\reportnumber{}

\input{texts/00_abstract}

\begin{document}

\maketitle

\input{texts/01_introduction}
\input{texts/03_prelim}
\input{texts/04_diag}

\input{texts/05_train}
\input{texts/06_train_result}
\input{texts/02_relatedwork}
\input{texts/07_conclusion}
\input{texts/08_ai_usage}

\bibliography{iclr2027_conference}
\bibliographystyle{plainnat}

\clearpage
\appendix

\input{texts/99_apdx}

\end{document}

%% file: math_commands.tex
\usepackage{amsmath,amsfonts,bm}

\def\eqref#1{equation~\ref{#1}}

\def\1{\bm{1}}

\DeclareMathAlphabet{\mathsfit}{\encodingdefault}{\sfdefault}{m}{sl}
\SetMathAlphabet{\mathsfit}{bold}{\encodingdefault}{\sfdefault}{bx}{n}



%% file: texts/00_abstract.tex
\begin{abstract}
Scaling test-time computation is a powerful way to improve language-model reasoning, and is particularly appealing for small reasoning models (sRMs) that are cheap to serve. 
However, is additional thinking always the right operation?
By intervening at intermediate reasoning states across two model families and multiple scales, we find that self-refinement largely consolidates probability mass onto solutions already reachable from the current state, rather than making new ones reachable. 
These interventions reveal two failure regimes: \emph{execution bottlenecks}, where the correct path is reachable and reflection can recover it, and \emph{knowledge bottlenecks}, where relevant external information makes it reachable. 
Motivated by this distinction, we introduce \modelname-, a selective querying framework, and train 4B and 8B variants to reason first, diagnose what remains unresolved, and, at a knowledge bottleneck, query stronger models whose parametric knowledge extends beyond its own.
Supervised fine-tuning bootstraps a multi-depth query action, and cost-aware reinforcement learning calibrates whether to query, what to ask, and how much to spend.
On 1,158 hard problems across six benchmarks, \modelname/-4B achieves 45.96\% pass@8, surpassing Qwen3-14B (41.64\%) at 2.7$\times$ lower serving cost, while also exceeding Qwen3-8B in pass@1 (16.85\% vs.\ 15.31\%).
Scaling to \modelname/-8B further improves pass@8 to 51.81\%.
\end{abstract}

%% file: texts/01_introduction.tex
\input{figs_tex/Concept_mock}
\section{Introduction}

Scaling test-time computation has emerged as a powerful way to improve language-model reasoning, where longer reasoning traces, repeated sampling, and self-refinement can each improve performance on challenging problems~\citep{snell2024scaling,brown2024large,muennighoff2025s1,madaan2023selfrefine,wu2026reasoning}.
This paradigm is particularly appealing for small reasoning models (sRMs), which are cheap to serve~\citep{liu2024mobilellm} yet lag behind their larger counterparts, since it promises to close this gap by thinking longer rather than by growing larger.
However, more computation is not uniformly useful: its benefit varies widely across problems and reasoning states~\citep{snell2024scaling}, and simply further prompting a model to reconsider its reasoning does not reliably repair an incorrect solution~\citep{huang2024large,d2026illusion}.
This limitation is especially acute for sRMs, since their failures reflect not only weaker reasoning capacity and limited learnability from stronger teachers~\citep{li2025small} but also gaps in their parametric knowledge~\citep{calderon2026empty,kang2026t1}, which additional reflection alone cannot fill.

Failed reasoning trajectories arise from two different bottlenecks (\cref{fig:concept}).
In an \emph{execution bottleneck}, the correct solution remains reachable, and additional computation can recover it through verification, revision, or exploration~\citep{madaan2023selfrefine,kim2026understanding,setlur2026e3,wang2025beyond}.
In a \emph{knowledge bottleneck}, further reasoning over the same state is insufficient, while relevant external information can make a correct path reachable.
The central question is therefore not whether a model should \emph{think more}, but whether more thinking is the right operation at all.

This distinction becomes especially consequential when sRMs serve as \emph{cognitive cores} of larger reasoning systems, with external tools such as retrieval, code execution, or stronger models extending their capabilities~\citep{yao2023react,schick2023toolformer,gou2024tora,jin2025search,lin2025understanding}.
Recent work has argued that such agents should seek external help only when their epistemic needs cannot be resolved internally~\citep{wang2025toward}.
Existing systems can already learn when and how to invoke external help~\citep{jin2025search,su2025toolorchestra,zeng2026learning}, but generally optimize tool use directly, without explicitly diagnosing whether the current reasoning state remains internally recoverable or instead requires new information.
Our analysis makes this boundary operational at the level of intermediate reasoning states: first diagnose the bottleneck, then decide \emph{what} information to request and \emph{how much} external computation to spend obtaining it.
This raises our central question: \emph{Can a small model distinguish these bottlenecks from its evolving reasoning state and use that diagnosis to acquire the right information at the right level of external computation?}

We study this question through counterfactual interventions, in which we prompt further reflection or supply relevant information at intermediate reasoning states of eight reasoning models across two families and multiple scales.
Contrary to the natural hypothesis that sRMs simply fail to notice their own uncertainty, we find that they express it frequently but rarely turn it into progress: reflection mainly consolidates probability mass onto already reachable solutions, recovering execution bottlenecks but providing little benefit at knowledge bottlenecks.
Relevant external information, by contrast, makes these paths reachable, but larger models use it more effectively. sRMs thus face knowledge bottlenecks more often and benefit less from assistance. Seeking and using help must therefore be learned rather than merely prompted.
Effective test-time computation therefore requires choosing not only \emph{how much} to compute, but \emph{what kind} the current state requires.

Motivated by this finding, we introduce \modelname-, a framework that trains an sRM as a cognitive core that reasons first and selectively acquires missing knowledge from an external model when it reaches a knowledge bottleneck.
We expose external models as a \emph{multi-depth query} action within the reasoning trajectory, letting the model decide \emph{whether} to query, \emph{what} to ask, and \emph{how much} external computation to allocate across backends of increasing strength and cost. The external model sees only the query (not the problem), and the sRM resumes reasoning from the observation.
We bootstrap this behavior with supervised fine-tuning on a small set of rescue trajectories, where a single query turns a failed continuation into a success, then apply cost-aware reinforcement learning so the policy uses cheap parametric reasoning when sufficient and pays for help only when needed.

We validate \modelname- on 1,158 hard problems from six benchmarks spanning mathematics, science, medicine, and general reasoning. \modelname--4B, trained from Qwen3-4B, more than doubles the pass@8 of its base model (21.2\% to 46.0\%) and surpasses Qwen3-14B at 2.7$\times$ lower serving cost.
At 8B, it raises pass@8 from 34.2\% to 51.8\% without increasing serving cost.
It also outperforms alternatives that strengthen self-refinement, retrieve documents, or query a stronger model upfront, with the margin over the last widening on harder problems.
Further analyses show that reinforcement learning turns one-shot delegation into iterative information acquisition through repeated cheap queries, attaining the performance of the strongest backend at nearly the cost of the cheapest.


Our contributions are threefold.
(i) We diagnose reasoning failures and show that reflection mainly consolidates already reachable solutions rather than making new ones reachable.
(ii) We show that sRMs not only possess less parametric knowledge, but are also less able to exploit external help, making effective help-seeking itself a learned capability.
(iii) We train 4B and 8B models to reason first and selectively query stronger models, outperforming larger models at lower serving cost.

%% file: figs_tex/Concept_mock.tex
\begin{figure}[h]
    \vspace{0.1in}
    \centering
    \includegraphics[width=0.975\linewidth]{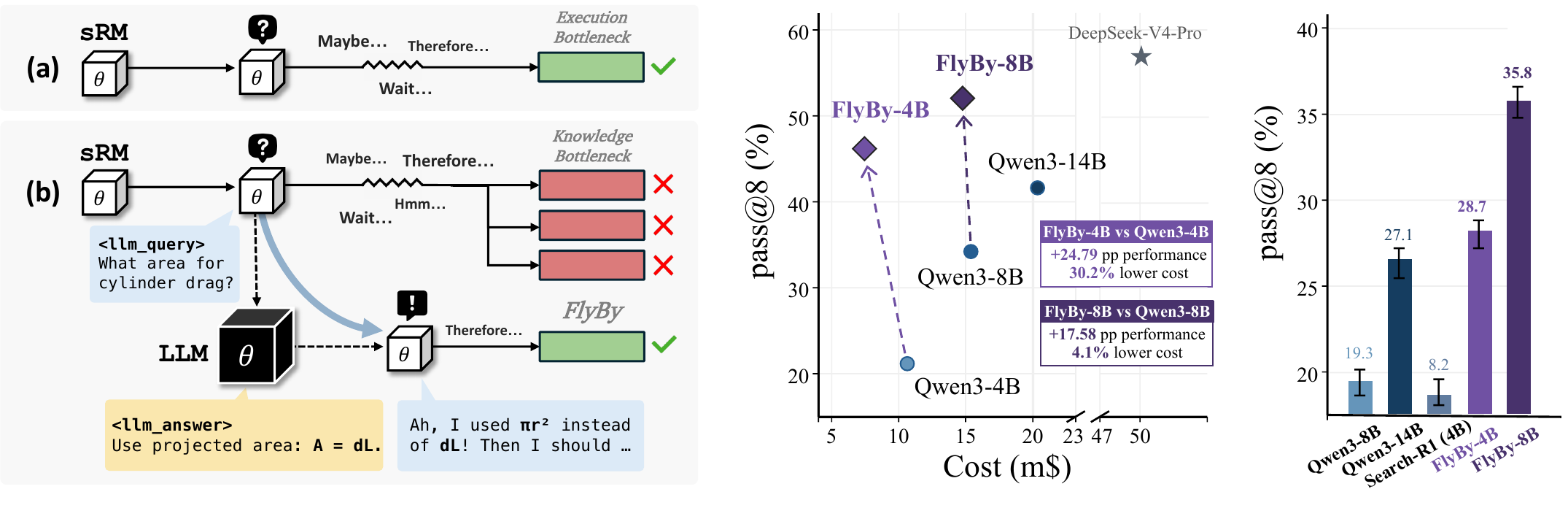}
    \caption{
        \small
        \textbf{Overview.}
        \textbf{Left:} sRM failures exhibit two bottlenecks: (a)
        \emph{execution bottlenecks}, where further reasoning can recover a reachable solution, and (b) \emph{knowledge bottlenecks}, where scarce parametric knowledge requires external information. \modelname. addresses both bottlenecks via local reasoning and selective querying; see \cref{fig:qualitative_phy} for an example.
        \textbf{Middle:} \modelname. improves the performance-cost trade-off, outperforming same-scale models at lower cost and surpassing Qwen3-14B with \modelname.-4B; see~\cref{tab:main_table} for details.
        \textbf{Right:} On problems unsolved by Qwen3-4B in 16 rollouts, \modelname.-4B reaches 28.7\% pass@8, surpassing Qwen3-14B.
    }
    \label{fig:concept}
    \vspace{-0.15in}
\end{figure}

%% file: texts/03_prelim.tex
\section{Preliminaries}
\label{sec:prelim}


Since our analysis intervenes at intermediate reasoning states, we first describe how we probe such states, how we locate where a model reflects, and how we measure the effect of intervening.

\paragraph{Reasoning states and state probing.}

Let $x\in\mathcal{D}$ be a problem with ground-truth answer $y^\star$.
A reasoning model $\pi$ generates a trace $z_{1:T}$ and terminates with a final answer $a$, and we call $s_t=(x,z_{<t})$ the reasoning state before $z_t$.
Since a single trace only shows whether the model happened to succeed, we instead probe a state $s$ by appending a cue $q$ and sampling $N$ continuations with final answers $a_q^{(1)}, \ldots, a_q^{(N)}$. The \emph{value} of $s$ is the fraction of continuations that reach the correct answer,
\begin{equation}
    V_q(s)=\frac{1}{N}\sum_{n=1}^N \mathbb{I}\left[a_q^{(n)}=y^\star\right],
    \label{eq:V_def}
\end{equation}
and the \emph{answer entropy} $\mathcal{H}_q(s)$ is the entropy of the resulting answer distribution, which measures how varied the reachable answers are.
For the null cue $q=\emptyset$, which leaves the state unmodified, we write $V(s)$ and $\mathcal{H}(s)$.
Each state thus lies on the $V\!\mathcal{H}$-plane, where productive reasoning moves toward $(V,\mathcal{H})=(1,0)$, at which the model is correct and settled on a single answer.

\paragraph{Epistemic verbalizations (EVs).}

To locate where a model attempts self-refinement, we use \emph{epistemic verbalizations} (EVs), short hedging or checking expressions with which a model pauses to reconsider its reasoning~\citep{kim2026understanding,wang2025beyond}.
We adopt the lexicon $\mathcal{E}$ of \citet{kim2026understanding} verbatim, nine expressions such as \texttt{wait}, \texttt{hmm}, and \texttt{alternatively} (full list in \S\ref{apdx:probe}) and count any case-insensitive whole-word match of $\mathcal{E}$ as an EV occurrence.
An EV is \emph{endogenous} if the model emits it on its own, and \emph{exogenous} if we insert it as a cue $q$ at a state of our choice.

\paragraph{Paired interventions.}

Because reasoning states differ widely in how promising they are, we measure each intervention against a counterfactual from the same state that differs only in the intervention. For an endogenous EV $e$ emitted at $s$, we compare continuing after $e$ against decoding from $s$ with $\mathcal{E}$ banned, and for an exogenous $q$, against the null cue:
\begin{equation}
    \Delta V(e)=V_e(s)-V_{\mathrm{noEV}}(s),
    \qquad
    \Delta V_q(s)=V_q(s)-V(s).
\end{equation}
The changes in answer entropy, $\Delta \mathcal{H}(e)$ and $\Delta \mathcal{H}_q(s)$, are defined analogously in \S\ref{apdx:probe}.

%% file: texts/04_diag.tex
\section{Understanding When Thinking Is Not Enough}
\label{sec:understanding}
Why do small reasoning models fail even when given sufficient opportunity to reason?
A natural hypothesis, motivated by prior work on self-refinement~\citep{kim2026understanding,d2026illusion,huang2024large}, is that small models are less capable of recognizing uncertainty and revising their intermediate reasoning.
Under this view, their performance gap arises from ineffective strategic allocation of reasoning effort: larger models can identify unproductive trajectories and redirect computation, whereas smaller models fail to do so.
Alternatively, progress may require information that cannot be reliably recovered from the current state through internal reasoning alone.

\paragraph{Experiment setup.}
\input{figs_tex/fig_diag1}
We conduct our analysis using Qwen3-0.6B,1.7B,4B,8B,14B~\citep{yang2025qwen3} and Gemma4-E2B,E4B,12B~\citep{team2026gemma}. 
For mathematical reasoning, we use 93 competition problems from \texttt{AIME25/26}~\citep{MAA_AIME} and \texttt{HMMT-Feb2026}~\citep{hmmt2026feb}, and for scientific reasoning, we use \texttt{GPQA-Diamond}~\citep{rein2023gpqa} in an open-ended setting. 
For each model and problem, we sample 16 independent rollouts with a maximum generation budget of 32K tokens using \texttt{vLLM}~\citep{kwon2025vllm}. 
We retain model-problem pairs with an empirical solve rate in $[0.25,0.75]$ to focus on problems that are nontrivial but still exhibit evidence of a viable solution path. 
From the retained rollouts, we identify 13.4K endogenous EV occurrences and apply the paired counterfactual described in \S\ref{sec:prelim}, using $N=8$ continuations per condition. 
This yields 215K counterfactual continuations, which together with 57K information-conditioned continuations give 272K continuations in total, each with a maximum generation budget of 32K tokens.

\subsection{Execution and knowledge bottlenecks}
To understand the source of reasoning failures, we distinguish two failure regimes based on whether a correct solution remains internally reachable, and then study how self-refinement and external information affect each regime.
We define an \emph{execution bottleneck} as a state from which a correct solution can be practically reached through the model's own reasoning, but the model cannot reliably realize it through its own reasoning process, and a \emph{knowledge bottleneck} as a state where the information needed for a correct solution cannot be reliably accessed, reconstructed, or utilized through further internal reasoning alone.
Since true reachability cannot be determined from finite samples, we operationalize this distinction using the estimated state value $V(s)$: states with at least one observed successful continuation ($V(s)>0$) are treated as \emph{execution-like}, whereas states with no observed successful continuation ($V(s)=0$) are treated as \emph{knowledge-like}.
This distinction is intended to capture practical accessibility under bounded reasoning rather than absolute reachability.
We next study how self-refinement and external information operate under these bottlenecks. 
Analysis on the sensitivity of this operational distinction to the continuation budget is in \S\ref{apdx:robust}.

\subsection{Self-refinement in small reasoning models}
\label{sec:ev}

We first ask whether small reasoning models fail because they do not recognize or express uncertainty.
Using the fixed EV lexicon $\mathcal{E}$ defined in \S\ref{sec:prelim}, we measure endogenous EV frequency and estimate the causal effect of each EV $e\in\mathcal{E}$ on value and answer diversity, quantified by $\Delta V(e)$ and $\Delta \mathcal{H}(e)$, respectively. For intervention, we uniformly sample four EV occurrences from each collected rollout. Exact formulas are provided in \S\ref{apdx:probe}.

As shown in \cref{fig:EVfreq}, EVs remain common even in smaller models, while their causal effect on value increases substantially with model scale~\citep{song2025mind}.
This suggests that the key limitation of small models is not expressing uncertainty or initiating reflection, but effectively converting reflection into progress.
Consistent with our distinction, we found that self-refinement mainly benefits from uncertain (\cref{fig:EVmech}) and execution-like states (\cref{fig:EVbottleneck}) increasing values while reducing answer entropy (\cref{fig:VHvectorfield_ev}). 

\begin{takeaway}
    \textbf{Takeaway~\thesubsection.}
    Self-refinement primarily helps realize solutions that are already internally reachable. 
\end{takeaway}

\subsection{Effect of external information}
\label{sec:info}

Previous analysis shows that self-refinement can recover incorrect trajectories when a successful continuation is already internally reachable.
This raises a natural question about states where such recovery remains difficult: does the model merely fail to trigger an effective refinement, or does progress require information that cannot be reliably recovered through further internal reasoning?
We distinguish these possibilities through controlled interventions.
If the former is the primary limitation, explicitly prompting further reflection should substantially improve the continuation.
If the latter is important, problem-relevant external information should provide a markedly larger benefit.
Following the intervention setup of \citet{kim2026understanding}, we take incorrect reasoning traces and intervene at relative positions $\alpha\in\{0.2,0.5,0.8,0.9\}$.
At each state $s$, we append a cue $q$ and measure its effect through $\Delta V_q(s)$ and $\Delta\mathcal{H}_q(s)$.
We compare three interventions.
An epistemic cue $q_{\mathrm{EV}}$ encourages further reflection using the prompts
\emph{``Wait, is that correct?''},
\emph{``Wait, let me double-check.''}, and
\emph{``Hmm, I'm not sure this is right.''},
following \citet{muennighoff2025s1}.
For each problem, we also construct an \emph{oracle information} cue $q_{\mathrm{info}}$ using DeepSeek-V4-Pro~\citep{xu2026deepseek}, which provides concise problem-relevant information without revealing the gold answer.
Finally, $q_{\mathrm{random}}$ uses oracle cues drawn from unrelated problems, preserving the presence and form of additional information while removing semantic relevance.
Details of oracle information generation, including the prompt used to elicit it, are provided in \S\ref{apdx:oracle_info}.

\input{figs_tex/fig_diag2}
\paragraph{Relevant information fills the knowledge gap.}
\input{figs_tex/fig_dyn}
As shown in \cref{fig:excue}, random cues provide little benefit, epistemic cues yield modest improvements, and relevant information produces substantially larger gains in value.
Simply asking the model to reconsider its reasoning therefore does not reliably recover knowledge-like states, while relevant external information often does.
The gain cannot be explained by longer generation or random cue insertion alone, since it depends strongly on the relevance of the supplied information.
Knowledge-like states are also substantially more prevalent in scientific reasoning than in mathematical reasoning (\cref{fig:domain}), consistent with the greater reliance of scientific problems on specific factual and domain knowledge.
Ultimately, self-refinement and relevant information are complementary.
External information can make a correct path accessible from a knowledge-like state, after which self-refinement can verify and consolidate it toward $(V,\mathcal{H})=(1,0)$ as illustrated in \cref{fig:VHvectorfield_ev,fig:VHvectorfield_info}.

\paragraph{Small models are limited on both sides.}
One might expect smaller models to benefit most from external information because their weaker parametric knowledge leaves greater headroom for improvement~\citep{calderon2026empty}.
However, \cref{fig:infoUtil} shows that information utilization generally improves with model scale.
Smaller models are less capable of incorporating a relevant hint into their ongoing reasoning.
The apparent drop at the largest Qwen and Gemma models is induced by changes in the residual problem set; when restricted to problems shared with the next smaller model, the increasing trend is preserved (empty markers in \cref{fig:infoUtil}).
Small models therefore face two limitations: they more often enter knowledge-like states, and they are less capable of exploiting external information once provided.

\begin{takeaway}
    \textbf{Takeaway~\thesubsection.}
    Utilizing external information expands what is reachable, while self-refinement consolidates solutions that are already within reach. Yet small models are limited in both.
\end{takeaway}

%% file: figs_tex/fig_diag1.tex
\begin{figure}[t]
    \vspace{-0.1in}
    \centering
    \begin{subfigure}[b]{0.31\textwidth}
        \includegraphics[width=\linewidth]{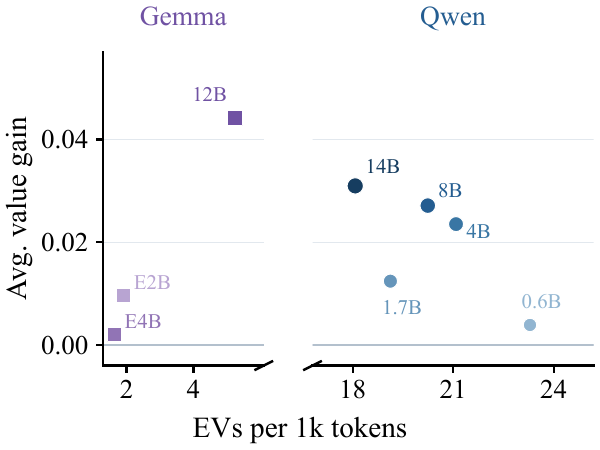}
        \caption{Frequency vs. effectiveness}\label{fig:EVfreq}
    \end{subfigure}\hfill
    \begin{subfigure}[b]{0.31\textwidth}
        \includegraphics[width=\linewidth]{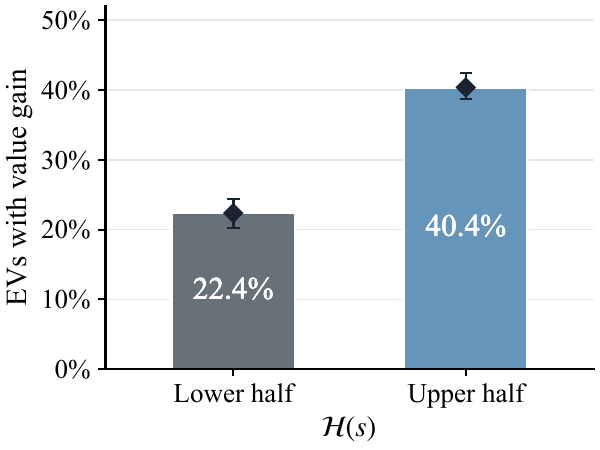}
        \caption{Uncertainty window}\label{fig:EVmech}
    \end{subfigure}\hfill
    \begin{subfigure}[b]{0.31\textwidth}
        \includegraphics[width=\linewidth]{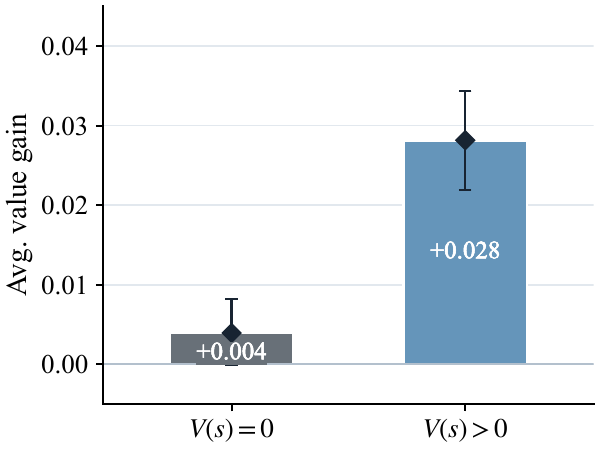}
        \caption{Recoverability analysis}\label{fig:EVbottleneck}
    \end{subfigure}\hfill
    \vspace{-0.05in}
    \caption{\textbf{Limits of self-refinement.} 
    (a) Higher EV frequency does not directly translate into larger value gains.
    (b,~c) Self-refinement is most effective in uncertain states where a correct solution remains reachable, rather than states requiring new information. $s$ is the state just before emitting $e$.
    }
    \label{fig:diag1}
    \vspace{-0.1in}
\end{figure}

%% file: figs_tex/fig_diag2.tex
\begin{figure}[t]
    \vspace{-0.15in}
    \centering
    \begin{subfigure}[b]{0.31\textwidth}
        \includegraphics[width=\linewidth]{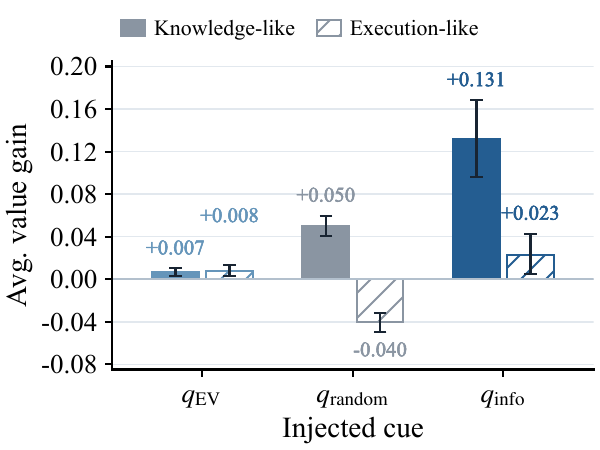}
        \caption{Effect of external cues}\label{fig:excue}
    \end{subfigure}\hfill
    \begin{subfigure}[b]{0.31\textwidth}
        \includegraphics[width=\linewidth]{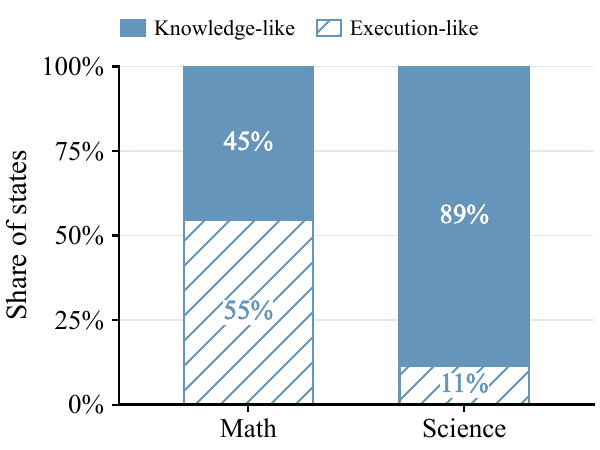}
        \caption{Bottlenecks by domain}\label{fig:domain}
    \end{subfigure}\hfill
    \begin{subfigure}[b]{0.31\textwidth}
        \includegraphics[width=\linewidth]{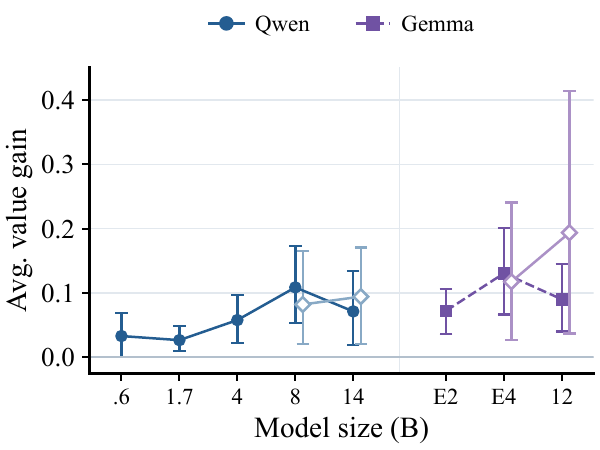}
        \caption{Information utilization}\label{fig:infoUtil}
    \end{subfigure}
    \vspace{-0.05in}
    \caption{\textbf{Diagnosing execution- and knowledge-like bottlenecks.}
    (a) On knowledge bottleneck states from incorrect trajectories, injecting oracle information yields a substantially larger value gain than an EV or random cue.
    (b) Among unresolved states, knowledge-like bottlenecks are substantially more prevalent in science than in math.
    (c) The benefit of relevant information increases with model scale.
    Blurred line indicates the performance on the shared problems.
    }
    \label{fig:diag2}
    \vspace{-0.15in}
\end{figure}

%% file: figs_tex/fig_dyn.tex
\begin{wrapfigure}{r}{0.56\textwidth}
    \centering
    \vspace{-0.2in}
    \captionsetup{font=small,skip=4pt}
    \captionsetup[subfigure]{
        font=small,skip=2pt,justification=centering
    }
    \begin{subfigure}[t]{0.485\linewidth}
        \centering
        \includegraphics[width=\linewidth]{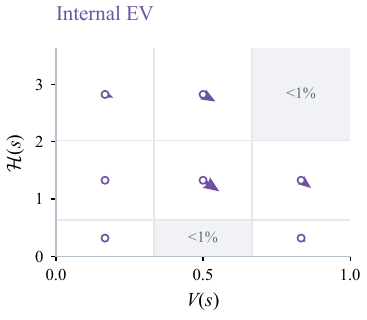}
        \caption{Internal EV dynamics}
        \label{fig:VHvectorfield_ev}
    \end{subfigure}\hfill
    \begin{subfigure}[t]{0.485\linewidth}
        \centering
        \includegraphics[width=\linewidth]{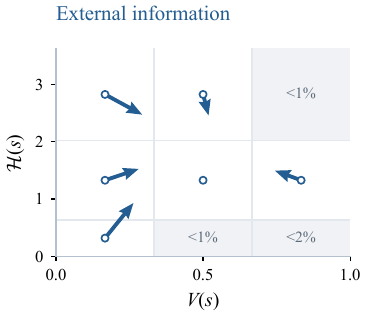}
        \caption{Information dynamics}
        \label{fig:VHvectorfield_info}
    \end{subfigure}
    \caption{\textbf{Reasoning-state dynamics.}
    (a, b) Average value-entropy transitions induced by internal EVs and external information, respectively. $s$ denotes the state right before intervention and arrow denotes the state transition.
    }
    \vspace{-0.1in}
    \label{fig:dyn}
\end{wrapfigure}

%% file: texts/05_train.tex
\section{Teaching Small Models When Thinking Is Not Enough}
\label{sec:train}


\subsection{Training Setup}

Motivated by the distinction above, we train \modelname--4B from Qwen3-4B to treat external reasoning as a \emph{selective querying} problem.
At each point in its reasoning trajectory, the model may either continue reasoning locally or query an external model, jointly choosing what information to request and how much external computation to invoke.
We use a two-stage pipeline: supervised fine-tuning (SFT) first bootstraps the query action space, followed by 80 steps of reinforcement learning (RL) to calibrate when querying is worthwhile and how much computation to allocate.
\paragraph{Tool design.}
\input{tables/tool_table}
We expose external models~\citep{xu2026deepseek,deepseekai2025deepseekv32} as a multi-depth query tool within the reasoning trajectory.
At each query step, the model decides how much compute to acquire by selecting a depth $d\in\{1,2,3\}$, receives the resulting observation, and resumes its own reasoning.
As summarized in \cref{tab:tool}, larger depths provide stronger external models and longer responses at higher cost.
To prevent trivial answer delegation, external models do not observe the original problem, and we reject queries with excessive $n$-gram overlap with the problem during both training and evaluation.

\paragraph{Cost.}
Following \citet{su2025toolorchestra}, we measure serving cost as the sum of local GPU inference cost and external API cost.
We convert both into USD using measured model throughput and third-party GPU/API prices from OpenRouter and Hyperbolic~\citep{openrouter,hyperbolic}, with a complete explanation of the cost model and the prices used provided in \S\ref{apdx:cost}.

\subsection{Two-stage optimization}

\paragraph{Bootstrapping via SFT.}

Directly learning selective querying through RL is difficult because the pretrained model has not learned to emit query actions or integrate external observations into its reasoning.
We therefore bootstrap this action space with a small, carefully curated SFT dataset.

Starting with failed trajectories from Qwen3-4B on \texttt{ArXivMath-Training}~\citep{dekoninck2026matharena} and \texttt{SuperGPQA}~\citep{pteam2025supergpqascalingllmevaluation}, we synthesize query-augmented trajectories in which external information reliably rescues an otherwise unsuccessful reasoning process.
We additionally construct targeted supervision for query generation and post-observation integration, and mix these examples with general reasoning trajectories from \texttt{OpenThoughts3-1.2M}~\citep{guha2025openthoughtsdatarecipesreasoning}.
Full details of trajectory synthesis, filtering, and SFT data curation are provided in~\S\ref{apdx:sft}.

\paragraph{Calibration via RL.}

\label{sec:rl}

The SFT model can emit query actions, but it has only imitated a fixed set of synthesized trajectories and has not learned when querying is actually worth its cost.
We therefore further optimize it with RL on problems drawn from \texttt{DAPO-17K-processed}~\citep{yu2025dapo}, \texttt{ArXivMath-Training}~\citep{dekoninck2026matharena}, the STEM subset of \texttt{GooseReason-0.7M}~\citep{lu2026goldengoose}, and \texttt{SuperGPQA}~\citep{pteam2025supergpqascalingllmevaluation}, excluding all problems used for SFT synthesis, which keeps the two training stages disjoint.

To make the policy cost-aware, we use a modified GRPO objective~\citep{shao2024deepseekmath} in which, following~\citet{su2025toolorchestra}, cost is penalized only for successful trajectories:
\begin{equation}
r(x,y)
=
\mathbb{I}[y=y^\star]
\left(1-\lambda \hat{C}(y)\right),
\end{equation}
where $\hat{C}(y)$ is the problem-level normalized cost of rollout $y$.
This prevents the policy from being rewarded for simply failing cheaply.
Importantly, querying can also be beneficial when external information substitutes for costly internal reasoning, such as retrieving a theorem instead of deriving it from first principles.
The objective therefore encourages the policy to acquire external information whenever doing so reduces the overall cost of reaching a correct solution.
We adopt DAPO-style decoupled clipping~\citep{yu2025dapo}, omit standard deviation normalization following~\citet{liu2025understanding}, and redact gold-answer spans from tool observations during training to prevent direct answer leakage.
Full details on the dataset mixture and training objective are in~\S\ref{apdx:rl}.

%% file: tables/tool_table.tex
\begin{wraptable}{r}{0.43\linewidth}
\vspace{-1.3em}
\centering
\caption{\textbf{Multi-depth query tool.}
API prices are USD per 1M tokens.}
\label{tab:tool}
\vspace{-0.4em}

\scriptsize
\setlength{\tabcolsep}{3pt}
\renewcommand{\arraystretch}{1.12}

\resizebox{\linewidth}{!}{%
\begin{tabular}{@{}clcc@{}}
\toprule
\textbf{$d$}
& \textbf{Backend}
& \textbf{Max tokens}
& \textbf{Input / Output} \\
\midrule
1 & DeepSeek-V4-Flash & 128   & 0.094 / 0.188 \\
2 & DeepSeek-V3.2     & 512   & 0.269 / 0.400 \\
3 & DeepSeek-V4-Pro   & 1,536 & 0.435 / 0.870 \\
\bottomrule
\end{tabular}
}

\vspace{-0.6em}
\end{wraptable}

%% file: texts/06_train_result.tex
\section{Experiments}
\input{tables/result_table}
\input{figs_tex/fig_rl1}
\subsection{Experiment setup}

\paragraph{Benchmarks and metric.}
We evaluate on six challenging reasoning benchmarks spanning mathematics, science, general knowledge, and medicine: \texttt{ArXivMath}~\citep{dekoninck2026matharena}, \texttt{GPQA-Diamond}~\citep{rein2023gpqa}, \texttt{SuperGPQA}~\citep{pteam2025supergpqascalingllmevaluation}, \texttt{ChemBench}~\citep{Mirza2025}, \texttt{MMLU-Pro}~\citep{wang2024mmlu}, and \texttt{MedXpertQA}~\citep{zuo2025medxpertqa}.
Our evaluation consists of 1,158 \emph{hard problems} across these benchmarks, defined as problems where vanilla Qwen3-4B achieves $\mathrm{pass@}1 \leq 0.25$.
Our primary metric is hard-problem $\mathrm{pass@}8$, reflecting coverage of challenging problems, while serving cost is defined in~\S\ref{sec:train}.
Both metrics are estimated from 16 rollouts; benchmark and filtering details are provided in~\S\ref{apdx:bench}.

\paragraph{Baselines and our method.}
We evaluate six approaches:
(1) \textbf{Larger Models}: Qwen3-8,14B and full delegation to DeepSeek-V4-Pro;
(2) \textbf{ForkingRL}, which post-trains Qwen3-4B on high-entropy forking tokens~\citep{wang2025beyond};
(3) \textbf{Search-R1}, which trains Qwen3-4B to retrieve documents during reasoning~\citep{jin2025search};
(4) \textbf{Query Opening}, which queries DeepSeek-V4-Flash once before reasoning;
(5) \textbf{Prompt Only}, which provides only the tool interface; and
(6) \textbf{\modelname- (Ours)}: \modelname--4B and \modelname--8B, trained from Qwen3-4B and Qwen3-8B, respectively, with \modelname--4B-SFT and vanilla Qwen3-4B as the SFT-only and base-model references.

\subsection{Main Results}
\input{figs_tex/fig_rl2}

\paragraph{Improved accuracy and coverage.}
As shown in \cref{tab:main_table}, light SFT achieves 36.39\% pass@8, and cost-aware RL further improves it to 45.96\% with only a marginal increase in serving cost, outperforming Qwen3-14B at 2.7$\times$ lower cost. On problems unsolved by Qwen3-4B in 16 no-tool rollouts, \modelname--4B achieves 28.7\% pass@8, extending coverage beyond the observed successes of the base model (\cref{fig:reachability}). Importantly, this improvement is not limited to multi-rollout coverage.
\modelname--4B also outperforms Qwen3-8B and Query Opening at pass@1 (\cref{tab:main_table_pass1}), indicating that RL improves the quality of individual reasoning trajectories rather than merely increasing the chance of obtaining a successful trajectory across repeated sampling.
Full pass@1 results are in~\S\ref{apdx:eval}.

\paragraph{Targeted information acquisition.}
ForkingRL provides little improvement over the base model, consistent with our finding that additional internal self-refinement is ineffective when hard problems are dominated by knowledge bottlenecks.
Search-R1 improves coverage through retrieval, but remains substantially below our model-based querying despite returning much longer contexts per call (2,015 characters on average).
In contrast, \modelname/-4B uses short external responses (105 output tokens on average) over multiple targeted queries, suggesting that resolving the current reasoning bottleneck matters more than simply supplying more external text. 

Query Opening is considerably stronger than the other baselines, but \cref{fig:cgain} shows that its gap to our adaptive policy widens as problems become harder.
We attribute this widening gap to the diminishing utility of a broad, upfront query without sufficient problem-specific reasoning. As problem difficulty increases, a generic request is less likely to surface the particular knowledge needed to resolve the failure.
This suggests that the benefit of adaptive querying comes not merely from access to a stronger model, but from first reasoning about the problem to identify what information is missing and only then acquiring targeted external assistance.

To test whether \modelname/ queries when further reasoning is unlikely to help, we sample 50 \texttt{SuperGPQA} problems and generate 8 rollouts per problem with \modelname/-4B.
At each query state $s$, we branch into two counterfactuals: execute the selected query, or suppress it and apply budget forcing~\citep{muennighoff2025s1} until the model attempts another query or generates 512 additional tokens, and probe all states using Qwen3-4B.
Querying increases state value by 5.30 pp on average, whereas additional reasoning yields essentially no improvement (\cref{fig:query_gain}), indicating that the learned policy tends to query precisely where further internal reasoning is ineffective.

\paragraph{Economic analysis.}
Although \modelname{}-4B is trained under fixed GPU and API prices, its cost-performance advantage remains robust to substantial price variation.
In \cref{fig:price}, we sweep GPU and API prices and partition the resulting price plane by the configuration achieving the highest pass@8 per USD.
\modelname{}-4B remains optimal across a broad range of price regimes, indicating that its economic advantage is not tied to a particular pricing assumption.

\paragraph{Effect of reinforcement learning.}

Compared with its SFT initialization, \modelname/-4B maintains substantial query probability at later turns (\cref{fig:queryprob}) indicating that RL develops repeated querying beyond the single-query behavior demonstrated during SFT.
Additionally, \modelname/-4B achieves coverage close to fixed depth-3 querying at a cost close to fixed depth-1 querying (\cref{fig:fixdepth}).
More analysis on RL is provided in~\S\ref{apdx:rl_analysis}.

\paragraph{Scaling to a stronger cognitive core.}
We further trained \modelname/-8B based on Qwen3-8B.
As demonstrated in \cref{tab:main_table}, \modelname/-8B reaches 51.81\% pass@8, improving over \modelname/-4B by 5.85 percentage points and outperforming Qwen3-14B by 10.17 points while remaining cheaper to serve.
This scaling trend is consistent with our analysis, as larger models not only possess greater parametric knowledge but also benefit more from utilizing given information (\cref{fig:infoUtil}).

Further experiments on backend generalization, tool leakage and price fluctuations are provided in \S\ref{apdx:ae}, with qualitative examples of mitigating knowledge bottlenecks and the corresponding dynamics on the $V\mathcal{H}$ plane in~\S\ref{apdx:ex}.

%% file: tables/result_table.tex
\begin{table}[t]

\vspace{-0.05in}
\centering

\caption{
\textbf{Main results.} Performance on hard reasoning problems across six benchmarks.
We report pass@8 and inference cost per eight rollouts in milli-dollars (m\$, $10^{-3}$ USD). Best and second-best results are \textbf{bolded} and \underline{underlined}, respectively. Frontier-scale DeepSeek-V4-Pro is excluded from the ranking. For Search-R1 baseline, we exclude retrieval costs due to ambiguity.
}

\label{tab:main_table}

\vspace{-0.08in}

\small

\setlength{\aboverulesep}{0pt}
\setlength{\belowrulesep}{0pt}
\renewcommand{\arraystretch}{1.15}
\newcommand{\gr}[1]{\textcolor{gray}{#1}}

\resizebox{\textwidth}{!}{
\setlength{\tabcolsep}{3.8pt}

\begin{tabular}{l cccccc cc}
\toprule

\addlinespace[2pt]

\multicolumn{1}{c}{\textbf{Model}}
& \textbf{ArXivMath}
& \textbf{GPQA-D}
& \textbf{SuperGPQA}
& \textbf{ChemBench}
& \textbf{MedXpertQA}
& \textbf{MMLU-Pro}
& \textbf{Avg Perf.}
& \textbf{Avg Cost ($\downarrow$)} \\

\addlinespace[1pt]
\midrule
\addlinespace[1pt]

Qwen3-4B
& \phantom{0}8.34
& 35.54
& 24.17
& 25.25
& 15.66
& 18.07
& 21.17
& 10.63 \\

Qwen3-8B
& 10.69
& 49.40
& 35.30
& 45.99
& 26.26
& 37.75
& 34.23
& 15.38 \\

Qwen3-14B
& 14.46
& 52.63
& 43.85
& 59.30
& 32.82
& \underline{46.78}
& 41.64
& 20.35 \\

\addlinespace[1pt]
\midrule
\addlinespace[2pt]

ForkingRL
& \phantom{0}8.81
& 34.86
& 23.75
& 27.90
& 16.98
& 17.00
& 21.55
& 11.47 \\

Search-R1
& 11.17
& 36.38
& 27.14
& 38.48
& 15.53
& 26.17
& 25.81
& \phantom{0}8.35 \\

Query Opening
& 15.70
& 44.47
& 39.01
& 50.73
& 23.45
& 30.99
& 34.06
& \phantom{0}8.56 \\

Prompt Only
& 12.97
& 41.77
& 25.58
& 27.93
& 16.10
& 21.83
& 24.36
& \phantom{0}\textbf{6.68} \\

\rowcolor{citeblue!10}
\textbf{\modelname/-4B-SFT}
& 18.19
& 43.78
& 38.22
& 57.37
& 27.21
& 33.55
& 36.39
& \phantom{0}\underline{7.15} \\

\rowcolor{citeblue!20}
\textbf{\modelname/-4B}
& \textbf{21.78}
& \underline{60.01}
& \underline{45.65}
& \underline{70.73}
& \underline{35.37}
& 42.21
& \underline{45.96}
& \phantom{0}7.42 \\


\rowcolor{citeblue!30}
\textbf{\modelname/-8B}
& \underline{18.97}
& \textbf{66.95}
& \textbf{49.11}
& \textbf{81.32}
& \textbf{43.76}
& \textbf{50.73}
& \textbf{51.81}
& 14.75 \\

\midrule
\addlinespace[1pt]

\rowcolor{gray!8}
\dgr{DS-V4-Pro}
& \dgr{17.81}
& \dgr{76.91}
& \dgr{48.69}
& \dgr{73.52}
& \dgr{61.99}
& \dgr{62.43}
& \dgr{56.89}
& \dgr{50.08} \\

\bottomrule
\end{tabular}
}

\end{table}

%% file: figs_tex/fig_rl1.tex
\begin{figure}[t]
    \centering
    \begin{subfigure}[b]{0.384\textwidth}
        \includegraphics[width=\linewidth]{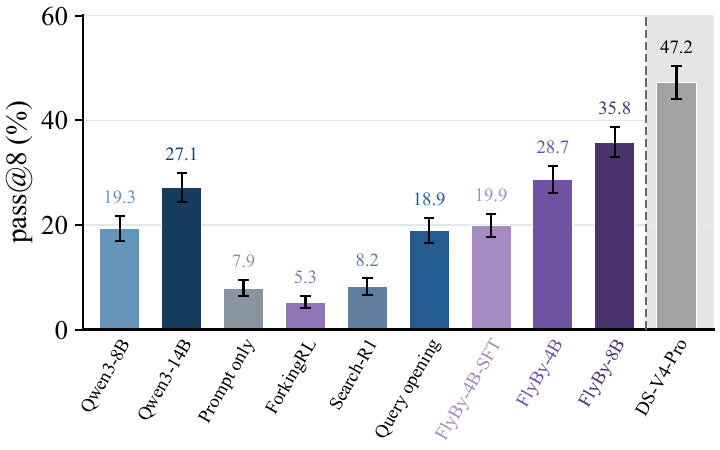}
        \caption{Reachability beyond Qwen3-4B}
        \label{fig:reachability}
    \end{subfigure}\hfill
    \begin{subfigure}[b]{0.32\textwidth}
        \includegraphics[width=\linewidth]{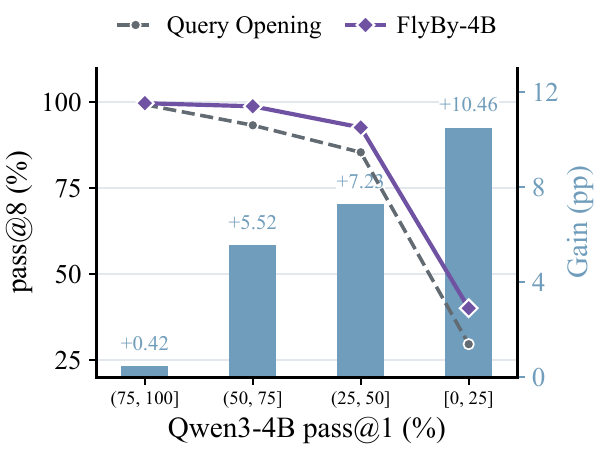}
        \caption{Gain over query opening}\label{fig:cgain}
    \end{subfigure}\hfill
    \begin{subfigure}[b]{0.256\textwidth}
        \includegraphics[width=\linewidth]{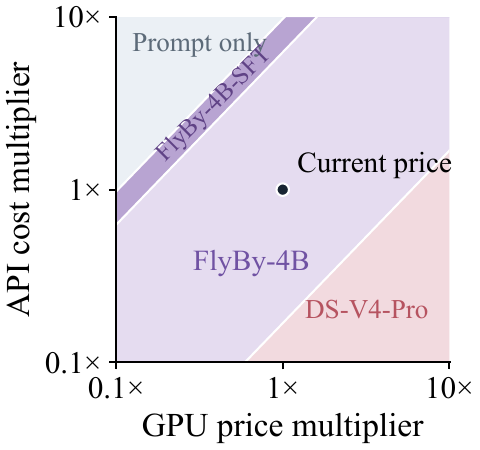}
        \caption{Economic analysis}\label{fig:price}
    \end{subfigure}
    \vspace{-0.05in}
    \caption{
    \textbf{Analysis of the learned querying policy.}
    (a) On problems with zero Qwen3-4B successes across 16 rollouts, \modelname/-4B reaches 28.7\% pass@8, expanding beyond the base model's observed reach.
    (b) Coverage gains over Query Opening are larger on harder problems.
    (c) \modelname{}-4B offers the best cost efficiency across most evaluated GPU and API price configurations.
    }
    \label{fig:rl1}
    \vspace{-0.1in}
\end{figure}

%% file: figs_tex/fig_rl2.tex
\begin{figure}[t]
    \vspace{-0.1in}
    \centering
    \begin{subfigure}[b]{0.31\textwidth}
        \includegraphics[width=\linewidth]{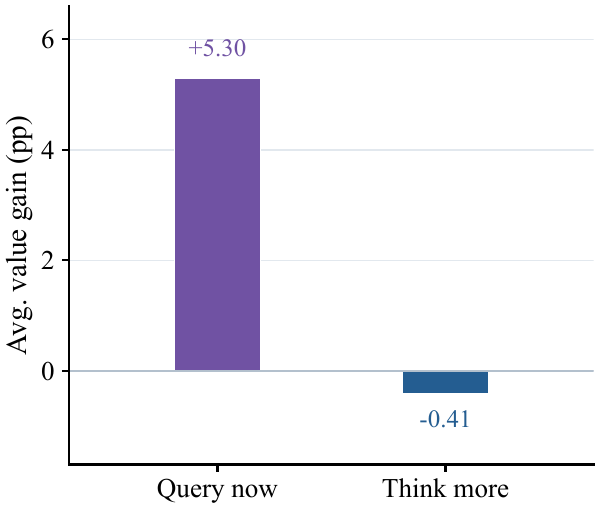}
        \caption{Recoverability}
        \label{fig:query_gain}
    \end{subfigure}\hfill
    \begin{subfigure}[b]{0.31\textwidth}
        \includegraphics[width=\linewidth]{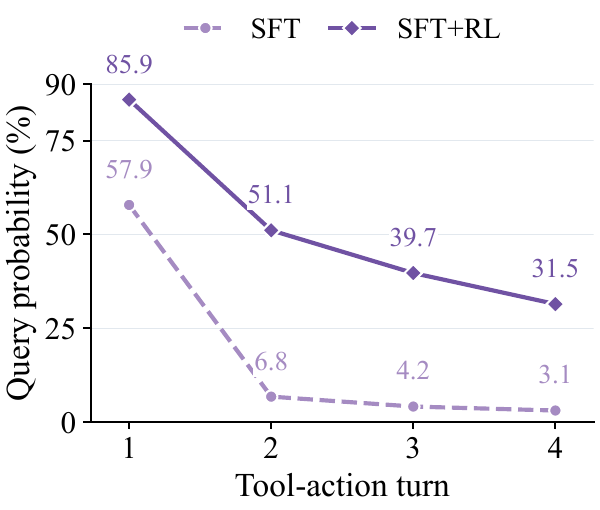}
        \caption{Iterative querying}
        \label{fig:queryprob}
    \end{subfigure}\hfill
    \begin{subfigure}[b]{0.31\textwidth}
        \includegraphics[width=\linewidth]{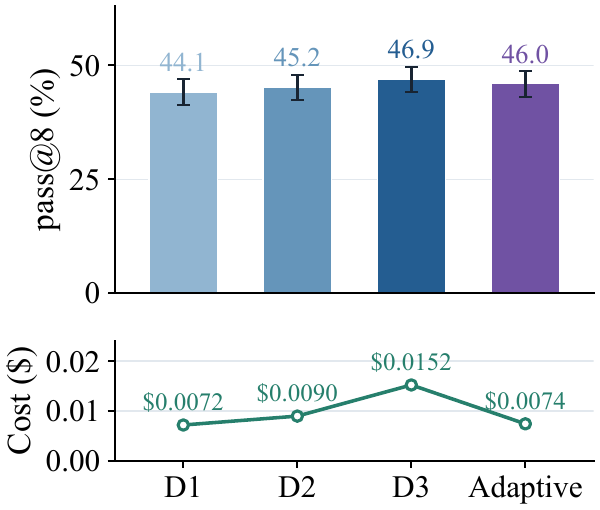}
        \caption{Fixed depth ablation}\label{fig:fixdepth}
    \end{subfigure}
    \vspace{-0.05in}
    \caption{
    \textbf{Effect of cost-aware RL.}
    (a) From the states where \modelname/-4B chooses to query, querying improves state value while further thinking does not.
    (b) Unlike SFT, RL maintains substantial query probability across later tool-action turns.
    (c) Fixing query depth for \modelname--4B shows that adaptive depth allocation attains near-deep-backend performance at near-shallow-backend cost.
    }
    \label{fig:rl2}
    \vspace{-0.1in}
\end{figure}

%% file: texts/02_relatedwork.tex
\section{Related Work}

\paragraph{Self-refinement via epistemic verbalization.}
Whether acquired during pretraining~\citep{liu2025understanding} or induced through reinforcement learning-based post-training~\citep{guo2025deepseek,shao2024deepseekmath}, self-refinement, often associated with the \textit{aha moment}~\citep{guo2025deepseek}, has emerged as an important characteristic of reasoning behavior in language models.
Recent studies~\citep{kim2026understanding,wang2025beyond} have characterized reasoning as a process of strategic information allocation under uncertainty, identifying token-level externalizations of uncertainty in the form of \textit{epistemic verbalizations} (EVs) or \textit{forking tokens}.
\citet{muennighoff2025s1} further showed that extending the reasoning process by appending \textit{``wait''} can improve model performance, while \citet{kim2026does} demonstrated that the loss of \textit{epistemic verbalizations} during self-distillation~\citep{hubotter2026reinforcement} degrades mathematical reasoning performance.
These findings suggest that recognizing uncertainty and reconsidering intermediate steps play an important role in effective reasoning.

\paragraph{Reasoning failures in small reasoning models.}
Reasoning performance generally improves with model scale, consistent with broader scaling trends observed in language models~\citep{kaplan2020scaling}.
However, the substantial inference cost of large reasoning models (LRMs) has motivated growing interest in developing capable small reasoning models (sRMs)~\citep{liu2024mobilellm}.
To reduce the capability gap induced by limited model capacity, prior work has explored distilling reasoning behaviors from larger teacher models into smaller students~\citep{agarwal2024policy,ko2024distillm,kang2025distillingllmagent}.
Nevertheless, \citet{li2025small} showed that small models exhibit not only lower reasoning performance but also a \textit{learnability gap}, which limits their ability to acquire reasoning behaviors from stronger teachers.
Moreover, \citet{calderon2026empty,kang2026t1} identified \textit{encoding failures} as a particularly prominent source of error in sRMs, suggesting that their failures can also arise from limitations in the information encoded in their parametric knowledge rather than from deficiencies in reasoning execution alone.

\paragraph{Adaptive tool use and model collaboration.}
Recent work trains language models to interleave reasoning with external tool use or stronger-model assistance, including dynamic sLM-LLM collaboration and cost-aware tool orchestration~\citep{su2025toolorchestra,zeng2026learning}.
These approaches establish that learned policies can adaptively decide when and how to seek help.
Our work asks a different question: \emph{which reasoning failures actually require external information?}
Through counterfactual interventions, we distinguish internally recoverable execution bottlenecks from knowledge bottlenecks, and use this distinction to motivate a policy that identifies the missing information and adaptively allocates the strength of external assistance.

%% file: texts/07_conclusion.tex
\section{Conclusion}
We studied when additional internal reasoning fails to help small reasoning models.
Our interventions reveal two failure regimes: \emph{execution bottlenecks}, where self-refinement can recover a reachable solution, and \emph{knowledge bottlenecks}, where external information is needed.
Building on this distinction, we train 4B and 8B models to selectively acquire external computation, learning whether, what, and how much to query.
The resulting \modelname--4B reaches 46.0\% pass@8 on hard problems while outperforming Qwen3-14B at 2.7$\times$ lower serving cost.
Our results suggest that effective test-time scaling requires knowing not only how to think longer, but when thinking is not enough.

\paragraph{Limitations and future directions.}
\modelname{} relies on external models, introducing practical concerns around availability, privacy, and reliability.
We discuss these limitations and promising future directions in~\S\ref{apdx:lim}.

%% file: texts/08_ai_usage.tex


%% file: texts/99_apdx.tex
\input{texts/apdx_texts/probe_detail}
\input{texts/apdx_texts/robust}

\input{texts/apdx_texts/oracle_information}
\input{texts/apdx_texts/state_dyn}
\input{texts/apdx_texts/training_details}
\input{texts/apdx_texts/evaluation_details}
\input{texts/apdx_texts/RL_analysis}
\input{texts/apdx_texts/additional_exp}
\input{texts/apdx_texts/examples}
\input{texts/apdx_texts/limit}

%% file: texts/apdx_texts/probe_detail.tex
\section{Probing Reasoning States}
\label{apdx:probe}

\paragraph{Probe policy.}
A reasoning state may be evaluated using a \emph{probe} policy
$\pi_{\mathrm{probe}}$ that need not equal the policy $\pi$ that produced the state.
Using a fixed probe policy places states generated by different policies on a common
comparison plane and keeps their trajectories comparable.
Unless stated otherwise, we use $\pi_{\mathrm{probe}}=\pi$ and suppress the probe policy
from the notation.

Given a state $s$, we append a short string cue $q$ and sample $N$ independent
continuations, writing $a_q^{(n)}$ for the final answer of the $n$-th continuation.
The null cue $q=\varnothing$ leaves the state unmodified.

\paragraph{Empirical value and answer entropy.}
Let $\mathcal{A}_q(s)$ be the set of distinct final answers observed across the $N$
continuations and let $K_q=\lvert\mathcal{A}_q(s)\rvert$.
We estimate the answer distribution as
\begin{equation}
\hat{p}_q(a\mid s)
=
\frac{1}{N}
\sum_{n=1}^{N}
\mathbb{I}\!\left[a_q^{(n)}=a\right].
\end{equation}
The empirical answer entropy is then
\begin{align}
\mathcal{H}_q(s)
&=
-\!\!\sum_{a\in\mathcal{A}_q(s)}\!\!
\hat{p}_q(a\mid s)\log \hat{p}_q(a\mid s)
+\frac{K_q-1}{2N},
\end{align}
where the second term is the Miller-Madow bias correction~\citep{miller1955note}.

\paragraph{EV lexicon.}
We use the EV lexicon of \citet{kim2026understanding} verbatim,
$\mathcal{E}=\{$ \texttt{wait}, \texttt{hmm}, \texttt{perhaps}, \texttt{maybe},
\texttt{actually}, \texttt{alternatively}, \texttt{seems}, \texttt{might}, \texttt{check} $\}$,
and count an EV occurrence as any case-insensitive whole-word match of an item of $\mathcal{E}$.

\paragraph{Suppressing reflection (from \S\ref{sec:prelim}).}
For an EV occurrence $e$, let $s$ denote the reasoning prefix immediately before $e$.
We write $V_e(s)$ and $\mathcal{H}_e(s)$ for the value and entropy obtained by appending
$e$ to $s$ and continuing generation.
As a paired counterfactual, $V_{\mathrm{noEV}}(s)$ and
$\mathcal{H}_{\mathrm{noEV}}(s)$ are obtained from the same prefix while decoding the
$N$ continuations with items in $\mathcal{E}$ banned.
We quantify the effect of the EV occurrence as
\begin{equation}
\Delta V(e)
=
V_e(s)-V_{\mathrm{noEV}}(s),
\qquad
\Delta\mathcal{H}(e)
=
\mathcal{H}_e(s)-\mathcal{H}_{\mathrm{noEV}}(s).
\end{equation}
\paragraph{Paired interventions.}
For a non-null cue $q$, its effect is measured relative to the null intervention on the
same reasoning state:
\begin{equation}
\Delta V_q(s)=V_q(s)-V(s),
\qquad
\Delta\mathcal{H}_q(s)=\mathcal{H}_q(s)-\mathcal{H}(s).
\end{equation}
Because both quantities are evaluated from the same underlying state, these differences
isolate the local effect of the intervention from variation across reasoning trajectories.

%% file: texts/apdx_texts/robust.tex
\newpage
\section{Robustness to the Continuation Budget}
\label{apdx:robust}
\input{figs_tex/fig_robust}

A natural concern is that states with no observed successful
continuations under $N=8$ may contain rare successful trajectories.
We therefore select 20 failed trajectories per domain for Qwen3-4B, evaluate four prefix states from each, and draw 64 new null continuations per state, yielding 160 states.
We recompute state classifications using the first $N\in\{8,16,32,64\}$ samples and denote $V_N(s)$ as an estimated value of state $s$ using $N$ continuations.

As expected, larger budgets reveal additional successes.
Among the 76 states with no success in the first eight new samples, 28 become positive by $N=64$ (\cref{fig:classification}).
Thus, zero observed success should be interpreted as a finite-sample criterion rather than literal unreachability.

Importantly, these newly positive states retain a large intervention gap (\cref{fig:excue_64}).
Hence, observing rare successful continuations does not imply that self-refinement can substantially amplify their probability.
This is consistent with our earlier finding that self-refinement primarily acts as a verification operation, consolidating probability mass around already plausible solutions rather than rescuing solutions with only vanishing endogenous probability.
External information, in contrast, can directly shift such low-reachability states by supplying information that is difficult to recover through further internal reasoning alone.

Our notion of a \emph{knowledge bottleneck} is therefore operational:
the relevant distinction is whether information is reliably accessible under bounded endogenous reasoning, not whether it is absolutely absent from the model.
The persistent intervention gap under higher-budget classification supports the robustness of our original $N=8$ criterion.

%% file: figs_tex/fig_robust.tex
\begin{wrapfigure}{r}{0.56\textwidth}
    \centering
    \vspace{-0.2in}
    \captionsetup{font=small,skip=4pt}
    \captionsetup[subfigure]{
    font=small,skip=2pt,justification=centering
    }
    \begin{subfigure}[t]{0.485\linewidth}
        \centering
        \includegraphics[width=\linewidth]{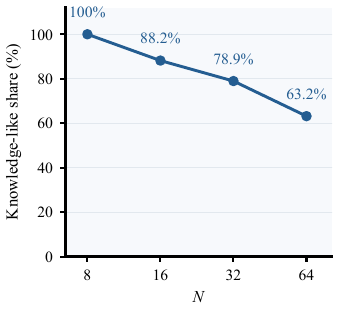}
        \caption{Classification results}
        \label{fig:classification}
    \end{subfigure}\hfill
    \begin{subfigure}[t]{0.485\linewidth}
        \centering
        \includegraphics[width=\linewidth]{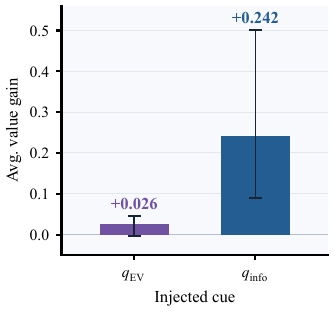}
        \caption{Effect on Value}
        \label{fig:excue_64}
    \end{subfigure}
    \caption{\textbf{Analysis on $N$.}
    (a) Some states classified as zero-success under \(N=8\) become positive with additional sampling.
    (b) Among these reclassified states, oracle information substantially increases continuation value, whereas epistemic-verbalization cues provide only limited gains.
    }
    \vspace{-0.1in}
    \label{fig:robust}
\end{wrapfigure}

%% file: texts/apdx_texts/oracle_information.tex
\section{Generating oracle information}
\label{apdx:oracle_info}

To examine the effect of injecting external information into incorrect reasoning trajectories, we construct problem-specific oracle information using DeepSeek-V4-Pro. We define oracle information as a minimal piece of missing knowledge or a problem-specific hint that can help a reasoning model recover from an incorrect trajectory without directly revealing the final answer.

For each problem, we prompt DeepSeek-V4-Pro to first solve the problem and then generate a targeted hint based on the resulting solution. The prompt explicitly instructs the model to avoid answer leakage, including the final answer itself, answer-equivalent expressions, and intermediate information that would make the answer trivially recoverable. For knowledge-intensive problems, the generated hint primarily provides missing domain knowledge, whereas for mathematical or symbolic problems, it describes the key reasoning strategy or derivation direction. 

We additionally screen generated hints for occurrences of the gold-answer string in the \texttt{idea} field. This flags four cases, all manually verified as incidental matches: three involve chemical indices or locants, and one refers to a rotation angle given in the problem. None of the flagged matches reveals the target answer.

%% file: texts/apdx_texts/state_dyn.tex



%% file: texts/apdx_texts/training_details.tex
\section{Training details}
\label{apdx:training}
\input{texts/apdx_texts/tool_detail}
\input{texts/apdx_texts/cost_detail}
\input{texts/apdx_texts/SFT_details}
\input{texts/apdx_texts/RL_details}
\input{texts/apdx_texts/hp_detail}
\input{texts/apdx_texts/8b_training}

%% file: texts/apdx_texts/tool_detail.tex
\subsection{Details on Tool}

We provide the exact prompts used for the tool interface introduced in \S\ref{sec:train}.

The system prompt used for our models is shown below:
\input{texts/prompts/ours_prompt}

The system prompt used for the LLM backend is shown below:
\input{texts/prompts/llm_sys_prompt}

For the $n$-gram filtering, we used $n=8$.
This constitutes a conservative leakage filter: prior work on benchmark decontamination uses exact $n$-gram matching with $n$ between 8 and 13, with 8 chosen as the minimum to avoid excessive spurious collisions at smaller $n$~\citep{brown2020language}. 

%% file: texts/prompts/ours_prompt.tex
\begin{promptbox}
\begin{Verbatim}[breaklines=true, breakanywhere=true]
You are a careful reasoning assistant. Solve the problem step by step.

Use this optional, expensive tool only for a specific external knowledge gap:

<llm_query depth="2">short question about the missing fact</llm_query>

Use exactly one double-quoted `depth` attribute and raw question text.
Do NOT use JSON or other attributes. LaTeX backslashes are literal, not
doubled or escaped.

Ask a SHORT question (under 300 characters) about a concept, theorem,
formula, or fact you are unsure of. Do NOT paste the problem; the assistant
cannot see it or solve it for you. Its reply appears as
<lrm_answer>...</lrm_answer>.

Normally one call is enough. After a reply, integrate it and finish.
Never repeat or rephrase a query; call again only for a distinct gap that
still blocks the answer. After a <tool_error>, do not repeat the same
malformed call.

`depth` buys the following oracle tier and response budget:
     depth 1: [DEPTH 1 LABEL] ([RELATIVE COST])
     depth 2: [DEPTH 2 LABEL] ([RELATIVE COST])
     depth 3: [DEPTH 3 LABEL] ([RELATIVE COST])

Calling costs you, and deeper calls cost much more. Ask only if the answer
would change what you do next, using the shallowest sufficient depth.

Finish with your final answer within \boxed{}.
\end{Verbatim}
\end{promptbox}

%% file: texts/prompts/llm_sys_prompt.tex
\begin{promptbox}
\begin{Verbatim}[breaklines=true, breakanywhere=true]
QUERY_CONTRACT = (
    "You are a knowledge assistant. A small model asks you a short question while solving a "
    "problem you CANNOT see. Answer with relevant concepts, theorems, formulas, or facts only. "
    "Do NOT attempt to solve any problem, do NOT give a final answer, a numeric result, or a "
    "multiple-choice letter."
)
\end{Verbatim}
\end{promptbox}

%% file: texts/apdx_texts/cost_detail.tex
\subsection{Details on Cost}
\input{tables/thruput_table}
\label{apdx:cost}

We measure inference cost by combining the local GPU cost incurred by the reasoning model with the API cost incurred by external tool calls. API prices are taken from OpenRouter~\citep{openrouter}, with the per-token input and output prices for each tool model reported in \cref{tab:tool}. For local inference, we use the hourly price of an NVIDIA H200 GPU from Hyperbolic~\citep{hyperbolic}, using a price snapshot collected on July 21, 2026, which gives $p^{\text{GPU/hr}}=\$3.49$.

To estimate local inference cost, we first measure the prefill and decoding throughput, in tokens per second, for each Qwen3 model on a single NVIDIA H200 GPU.
The measured throughput values are reported in~\cref{tab:throughput}.
These values are measured once before evaluation and kept fixed throughout all subsequent cost calculations.
For a trajectory $y$, we estimate its local inference cost as
\begin{equation}
\label{eq:local_cost}
\operatorname{Cost}_{\mathrm{local}}(y)
=
\frac{p^{\mathrm{GPU/hr}}}{3600}
\left(
\frac{T_{\mathrm{prefill}}}{v_{\mathrm{prefill}}}
+
\frac{T_{\mathrm{own}}}{v_{\mathrm{decode}}}
\right),
\end{equation}
where $T_{\mathrm{prefill}}$ denotes the number of tokens processed during prefill, $T_{\mathrm{own}}$ is the number of tokens generated by the local reasoning model, and $v_{\mathrm{prefill}}$ and $v_{\mathrm{decode}}$ denote the measured prefill and decoding throughput, respectively.

For external tool usage, we compute the API cost by summing the input and output token costs over all tool calls made during the trajectory:
\begin{equation}
\operatorname{Cost}_{\mathrm{API}}(y)
=
\sum_{c \in \mathcal{C}(y)}
\left(
T^{(c)}_{\mathrm{in}} p^{(c)}_{\mathrm{in}}
+
T^{(c)}_{\mathrm{out}} p^{(c)}_{\mathrm{out}}
\right),
\end{equation}
where $\mathcal{C}(y)$ is the set of tool calls in trajectory $y$, $T^{(c)}_{\mathrm{in}}$ and $T^{(c)}_{\mathrm{out}}$ are the corresponding input and output token counts, and $p^{(c)}_{\mathrm{in}}$ and $p^{(c)}_{\mathrm{out}}$ are their per-token API prices.

The total cost of a trajectory is then defined as
\begin{equation}
\operatorname{Cost}(y)
=
\operatorname{Cost}_{\mathrm{local}}(y)
+
\operatorname{Cost}_{\mathrm{API}}(y).
\end{equation}
This formulation allows us to compare methods using a common monetary cost measure that accounts for both local computation and externally acquired computation.

%% file: tables/thruput_table.tex
\begin{table}[t]
\centering
\caption{
\textbf{Measured inference throughput} for Qwen3 models on a single NVIDIA H200 GPU.
We report the prefill and decoding throughput used to estimate local inference cost in~\cref{eq:local_cost}.
}
\label{tab:throughput}

\small
\resizebox{0.62\textwidth}{!}{
\setlength{\tabcolsep}{10pt}
\begin{tabular}{l cc}
\toprule

\textbf{Model}
& \textbf{\shortstack{Prefill throughput\\(tokens/s)}}
& \textbf{\shortstack{Decode throughput\\(tokens/s)}} \\

\midrule

Qwen3-4B
& 66,199.94
& 5,166.24 \\

Qwen3-8B
& 40,114.82
& 4,056.16 \\

Qwen3-14B
& 22,564.44
& 2,823.72 \\

\bottomrule
\end{tabular}
}

\vspace{-0.1in}
\end{table}

%% file: texts/apdx_texts/SFT_details.tex
\subsection{Details on SFT}
\label{apdx:sft}
\input{texts/pseudo_codes/sft_traj}

\paragraph{High-precision SFT trajectory synthesis.}
Rather than collecting tool-use trajectories at scale, we construct a small set of trajectories in which external information reliably rescues a failed reasoning process.
We begin with failed no-tool rollouts from Qwen3-4B on 800 training problems, consisting of 400 problems from \texttt{ArXivMath-Training}~\citep{dekoninck2026matharena} and 400 from \texttt{SuperGPQA}~\citep{pteam2025supergpqascalingllmevaluation}.
For each failed rollout, we identify candidate intervention states around expressions of uncertainty or doubt, with fixed fractional positions used as fallbacks.
At each state, DeepSeek-V4-Pro generates a query from the problem and recent reasoning context, and the same query is evaluated with tools of increasing depth from \cref{tab:tool}.

We retain a trajectory only when tool access produces a clear and robust rescue.
Specifically, we compare four plain continuations with four tool-assisted continuations from the same prefix and require the selected state to have no successful plain continuation but at least three successful tool-assisted continuations.
Queries and observations are additionally filtered for validity, excessive problem overlap, and answer leakage.
The first correct continuation at the shallowest accepted tool depth is stored as the rescue trajectory.
This high-precision filtering reduces the original 800-problem pool to only 95 rescue trajectories from 95 distinct problems, including 37 from \texttt{ArXivMath-Training} and 58 from \texttt{SuperGPQA}.
\Cref{alg:sft-synthesis} provides the complete synthesis and filtering procedure.
We use only single-call rescues for SFT.

\paragraph{Curated supervision from rescue trajectories.}
Pilot experiments showed that full-trajectory supervision alone was insufficient to reliably induce tool use.
Because query tokens occupy only a small fraction of each trajectory, the resulting model rarely emitted queries.
Adding query-only supervision improved tool invocation, but often caused repeated queries after receiving an observation, suggesting that the model had not learned the transition from external information back to local reasoning.

We therefore decompose each rescue trajectory into three complementary supervision signals: (1) the full rescue trajectory, (2) a query example supervised only on the query action and its depth, and (3) an integration example supervised only on the first up to 128 policy tokens following the tool observation.
To preserve general reasoning ability, we additionally include verified correct no-tool trajectories and examples from \texttt{OpenThoughts3-1.2M}~\citep{guha2025openthoughtsdatarecipesreasoning}.
This follows the broader practice of combining general reasoning data with synthetic tool-use data, as in \texttt{ToolOrchestra}~\citep{su2025toolorchestra}, which incorporates \texttt{GeneralThought-430K}~\citep{lambert2025generalthoughtfiltered}.

The final SFT dataset contains five equally sized groups: 95 full rescue trajectories, 95 verified no-tool trajectories, 95 OpenThoughts reasoning examples, 95 query examples, and 95 integration examples.
This yields only 475 training examples in total, mixed at a $1{:}1{:}1{:}1{:}1$ ratio.
The query and integration groups are derived from the same 95 rescue trajectories rather than additional problem instances.
Prompts and external observations are masked from the loss, while selected policy tokens receive unit loss weight.

%% file: texts/pseudo_codes/sft_traj.tex
\begin{algorithm}[t]
\caption{SFT Trajectory Synthesis and Filtering}
\label{alg:sft-synthesis}
\small
\begin{algorithmic}[1]
\Require Failed no-tool trajectories $\mathcal{F}$; policy $\pi$;
query generator $G$; tools $\{\mathcal{T}_d\}_{d=1}^{3}$
\Ensure SFT dataset $\mathcal{D}_{\mathrm{query}}$

\State $\mathcal{D}_{\mathrm{query}} \gets \emptyset$
\State $\mathcal{C} \gets$ EV-position prefixes from $\mathcal{F}$,
paired with ground-truth answers

\ForAll{$(s_t, y^\star) \in \mathcal{C}$}
    \Comment{$s_t=(x,z_{1:t-1})$}
    \State Generate query $u_t \gets G(s_t)$; \textbf{continue} if invalid
    \State Sample $4$ no-tool continuations from $\pi(\cdot \mid s_t)$
    \State $b_0 \gets$ number of correct no-tool continuations

    \For{$d = 1,2,3$}
        \State Obtain and filter tool response
        $o_d \gets \mathcal{T}_d(u_t)$; \textbf{continue} if unavailable
        \State Sample $4$ continuations from
        $\pi(\cdot \mid s_t, u_t, o_d)$
        \State $b_d \gets$ number of correct tool-augmented continuations

        \If{$b_d \geq 2$ \textbf{and} $b_d - b_0 \geq 1$}
            \Comment{Collection criterion}
            \If{$b_0 = 0$ \textbf{and} $b_d \geq 3$}
                \Comment{SFT criterion}
                \State Add the first correct full trajectory to
                $\mathcal{D}_{\mathrm{query}}$
            \EndIf
            \State \textbf{break}
        \EndIf
    \EndFor
\EndFor

\State \Return $\mathcal{D}_{\mathrm{query}}$
\end{algorithmic}
\end{algorithm}

%% file: texts/apdx_texts/RL_details.tex
\subsection{Details on RL}
\label{apdx:rl}
\paragraph{Dataset Curation.}
We mix problems with Qwen3-4B no-tool solve probability in $[0.25,0.75]$ and problems likely to benefit from external information at approximately 6:4 ratio.
The former represent cases where the utility of querying is ambiguous, while the latter emphasize information-hard problems for which external computation has the greatest potential benefit.

\paragraph{Training Algorithm.}
As discussed in \S\ref{sec:rl}, we apply the cost penalty using problem-level normalized rollout costs.
For $N$ rollouts $\{y_i\}_{i=1}^{N}$ sampled for a problem $x$, we normalize the serving cost within each rollout group as
\begin{equation}
\hat{C}(y_i)
=
\frac{
\mathrm{Cost}(y_i) - \min_j \mathrm{Cost}(y_j)
}{
\max_j \mathrm{Cost}(y_j) - \min_j \mathrm{Cost}(y_j) + \varepsilon_c
},
\end{equation}
where $\varepsilon_c$ is a small constant for numerical stability.
The rollout reward is then computed using the cost-aware reward in \S\ref{sec:rl}.

Following Dr.~GRPO~\citep{liu2025understanding}, we center rewards within each rollout group without standard deviation normalization:
\begin{equation}
\hat{A}_i
=
r_i - \frac{1}{N}\sum_{j=1}^{N} r_j.
\end{equation}

We optimize the policy using the DAPO-style asymmetric clipped objective~\citep{yu2025dapo} with a KL constraint to the reference policy:
\begin{equation}
\label{eq:RL_token}
\begin{aligned}
\mathcal{L}_{\mathrm{RL}}(\theta)
=\\
-\mathbb{E}_{x \sim \mathcal{D},\, \{y_i\}_{i=1}^{G} \sim \pi_{\theta_{\mathrm{old}}}(\cdot \mid x)}
\Bigg[
\frac{1}{\sum_{i=1}^{G} |y_i|}
\sum_{i=1}^{G} \sum_{t=1}^{|y_i|}
\Big(
&\min\Big(
\rho_{i,t} \hat{A}_{i,t},\,
\mathrm{clip}\big(
\rho_{i,t},
1-\epsilon_{\mathrm{low}},
1+\epsilon_{\mathrm{high}}
\big)
\hat{A}_{i,t}
\Big) \\
&- \beta\, D_{\mathrm{KL}}^{(i,t)}
\Big)
\Bigg],
\end{aligned}
\end{equation}
where
\begin{equation}
\rho_{i,t}
=
\frac{\pi_\theta(y_{i,t} \mid x, y_{i,<t})}
{\pi_{\theta_{\mathrm{old}}}(y_{i,t} \mid x, y_{i,<t})},
\end{equation}
\begin{equation}
D_{\mathrm{KL}}^{(i,t)}
=
D_{\mathrm{KL}}\!\left(
\pi_\theta(\cdot \mid x, y_{i,<t})
\,\|\,
\pi_{\mathrm{ref}}(\cdot \mid x, y_{i,<t})
\right).
\end{equation}
and $\pi_{\mathrm{ref}}$ denotes the reference policy before RL, while $\beta$ controls the strength of the KL constraint. \paragraph{Training pipeline.}
We used \texttt{verl-tool}~\citep{jiang2025verltool} for RL training.

%% file: texts/apdx_texts/hp_detail.tex
\subsection{Hyperparameter}
Hyperparameters used throughout the training process are presented in \cref{tab:training-hparams}.
\input{tables/hp_table}

%% file: tables/hp_table.tex
\begin{table}[ht]
\caption{Hyperparameters for SFT and RL.}
\label{tab:training-hparams}
\centering
\small
\begin{tabular}{ll}
\toprule
\textbf{Parameter} & \textbf{Value} \\
\midrule
\multicolumn{2}{c}{\textbf{Supervised Fine-Tuning (SFT)}} \\
\midrule
Fine-tuning method & Full fine-tuning \\
Optimizer & AdamW \\
Learning rate & $2\times10^{-6}$ \\
LR scheduler & Cosine \\
Warmup steps & 8 \\
Weight decay & 0.01 \\
Max gradient norm & 1.0 \\
Training epochs & 2 \\
Effective batch size & 8 \\
Gradient accumulation steps & 8 \\
Max sequence length & 24{,}576 \\
Random seed & 42 \\
\midrule
\multicolumn{2}{c}{\textbf{Reinforcement Learning (RL)}} \\
\midrule
Fine-tuning method & Full fine-tuning \\
Optimizer & AdamW \\
Learning rate & $1\times10^{-6}$ \\
LR scheduler & Constant \\
Warmup steps & 0 \\
Weight decay & 0.01 \\
Max gradient norm & 1.0 \\
Training steps & 80 \\
Prompts per batch & 8 \\
Rollouts per prompt & 8 \\
PPO mini-batch size (sequences) & 64 \\
PPO epochs per rollout batch & 1 \\
Gradient accumulation steps & Dynamic \\
Max prompt length & 4{,}096 \\
Max policy-generated tokens & 16{,}384 \\
Max response length (incl.\ observations) & 22{,}592 \\
Temperature & 1.0 \\
Top-$p$ & 1.0 \\
Top-$k$ & $-1$ \\
KL loss coefficient & $1\times10^{-3}$ \\
Clipping $\epsilon_{\mathrm{low}}$ & 0.2 \\
Clipping $\epsilon_{\mathrm{high}}$ & 0.28 \\
Cost penalty $\lambda$ & 0.1 \\
Random seed & 42 \\
\bottomrule
\end{tabular}
\end{table}

%% file: texts/apdx_texts/8b_training.tex
\subsection{Training 8B Model}
\label{apdx:train_8b}

We trained \modelname/-8B from Qwen3-8B following the same overall training procedure as in~\S\ref{sec:train}, with two modifications to account for the stronger reasoning capability of the 8B backbone.
First, under the same single-query trajectory synthesis procedure (\cref{alg:sft-synthesis}), Qwen3-8B yielded only 87 retained trajectories, resulting in a smaller SFT dataset. We therefore increased the SFT learning rate to $3\times10^{-6}$.
Second, as Qwen3-8B can solve a larger fraction of problems without external assistance, we decreased the no-tool solvable examples in the RL mixture to 50\%.

%% file: texts/apdx_texts/evaluation_details.tex
\section{Evaluation details}
\label{apdx:eval}
\input{texts/apdx_texts/benchmark_details}
\input{texts/apdx_texts/baseline_details}
\input{texts/apdx_texts/eval_hp_details}
\input{texts/apdx_texts/pass1_detail}

%% file: texts/apdx_texts/benchmark_details.tex
\subsection{Details on Benchmarks}
\label{apdx:bench}

We provide details of the benchmarks used for evaluation in \cref{tab:benchmarks}.
Because the hard evaluation set is selected using an initial set of Qwen3-4B rollouts, we report Qwen3-4B performance using an independent set of fresh rollouts on the fixed evaluation set to avoid selection bias.
\input{tables/bench_table}

%% file: tables/bench_table.tex
\begin{table}[ht]
\centering
\caption{
\textbf{Details of the evaluation benchmarks.}
We report the task coverage, evaluation subset, and number of questions
before and after hard-subset selection.
The hard subset contains questions answered correctly by Qwen3-4B
in at most 4 of 16 rollouts and is fixed across all evaluated methods.
All random sampling uses seed 42.
}
\label{tab:benchmarks}

\small
\resizebox{\textwidth}{!}{
\setlength{\tabcolsep}{6pt}
\renewcommand{\arraystretch}{1.15}
\begin{tabular}{
    l
    >{\raggedright\arraybackslash}p{3.5cm}
    >{\raggedright\arraybackslash}p{6.5cm}
    rr
}
\toprule
\textbf{Benchmark}
& \textbf{Task coverage}
& \textbf{Evaluation subset}
& \textbf{\shortstack{Eval\\pool}}
& \textbf{Hard} \\
\midrule

ArXivMath
& Research-level mathematical reasoning from arXiv papers.
& All questions from the December 2025--June 2026 monthly releases,
pooled across months.
& 232 & 207 \\
\addlinespace

GPQA-Diamond
& Graduate-level reasoning in biology, chemistry, and physics.
& The complete Diamond subset.
& 198 & 80 \\
\addlinespace

SuperGPQA
& Graduate-level knowledge and reasoning across academic disciplines.
& A held-out sample of 500 questions from Science, Engineering,
Medicine, and Agronomy, stratified by discipline and difficulty
and disjoint from our training split.
& 500 & 252 \\
\addlinespace

MedXpertQA
& Expert-level medical knowledge and clinical reasoning.
& 500 questions randomly sampled from the 2,450-question
\texttt{Text/test} split.
& 500 & 410 \\
\addlinespace

MMLU-Pro
& Broad academic knowledge and reasoning across 14 subject areas.
& 500 questions randomly sampled from the official test split.
& 500 & 129 \\
\addlinespace

ChemBench
& Chemistry and materials science knowledge and reasoning.
& All single-answer multiple-choice questions from Organic Chemistry
(384), Materials Science (57), and Inorganic Chemistry (49),
pooled across the three domains.
& 490 & 80 \\

\midrule
\textbf{Total}
& & & \textbf{2,420} & \textbf{1,158} \\
\bottomrule
\end{tabular}
}

\end{table}

%% file: texts/apdx_texts/baseline_details.tex
\subsection{Details on baselines implementation}

\paragraph{ForkingRL~\citep{wang2025beyond}}
We follow the high-entropy token update strategy of ForkingRL, restricting the policy-gradient objective to the top 20\% of response tokens ranked by entropy, while retaining KL regularization over all response tokens.
We initialize the policy from Qwen3-4B and train it with GRPO using binary answer correctness as the reward.
We adapt the training data to our task setting and match the rollout and optimization budgets of our RL setup, including 8 prompts per batch, 8 rollouts per prompt, and a maximum of 16,384 policy-generated tokens per trajectory.
The model performs standalone reasoning without external tools during both training and evaluation.

\paragraph{Search-R1~\citep{jin2025search}}
We follow the Search-R1 framework for interleaving reasoning with retrieval through reinforcement learning.
Our implementation uses the canonical \texttt{Wiki-18} corpus and the \texttt{E5-base-v2} dense retriever, returning the top 3 passages for each search query.
We initialize the policy from Qwen3-4B and train it with GRPO, masking retrieved observations from the policy loss.
We adapt the training data and final-answer format to our evaluation pipeline and match the rollout and optimization budgets of our RL setup.
Each trajectory permits up to 4 search calls and 16,384 policy-generated tokens.
The reward is binary answer correctness, without penalties for retrieval frequency, latency, or monetary cost.

\paragraph{Query Opening}
We construct a training-free baseline that obtains external information before beginning its reasoning.
Given only the problem, the unmodified Qwen3-4B model generates one knowledge query with thinking disabled.
The query is submitted to the same depth-1 external model used by our method, with the same response budget.
Qwen3-4B then solves the problem with thinking enabled, conditioned on the problem, query, and returned response, without further tool calls.
Query generation and subsequent reasoning share a total budget of 16,384 policy-generated tokens.

%% file: texts/apdx_texts/eval_hp_details.tex
\subsection{Hyperparameters during evaluation}
During evaluation, we set the sampling temperature to $0.6$, top-$p$ to $0.95$, and top-$k$ to $20$.
All other hyperparameters are kept identical to those reported
in~\cref{tab:training-hparams}.

%% file: texts/apdx_texts/pass1_detail.tex
\subsection{Additional Results}
\input{tables/pass1_table}

We provide full pass@1 results on the hard subset in~\cref{tab:main_table_pass1}.
On hard problems, \modelname/-4B achieves 16.85\% pass@1,
outperforming Qwen3-8B (15.31\%) as well as all tool-use baselines.
It improves over Query Opening by 2.83 percentage points and
Search-R1 by 9.63 points, while requiring less than half the serving
cost of Qwen3-8B (0.93 vs.\ 1.92 m\$ per rollout).
Scaling to \modelname/-8B yields 22.19\% matching the performance of Qwen3-14B at $1.4\times$ lower serving cost.

%% file: tables/pass1_table.tex
\begin{table}[ht]

\centering

\caption{
    \textbf{Pass@1 results.} Performance on hard reasoning problems across six benchmarks. We report pass@1 and inference cost per rollout in milli-dollars (m\$, $10^{-3}$ USD). Best and second-best results are \textbf{bolded} and \underline{underlined}, respectively. Frontier-scale DeepSeek-V4-Pro is excluded from the ranking. For Search-R1 baseline, we exclude retrieval costs due to ambiguity.
}

\label{tab:main_table_pass1}

\vspace{-0.08in}

\small

\setlength{\aboverulesep}{0pt}
\setlength{\belowrulesep}{0pt}
\renewcommand{\arraystretch}{1.15}
\newcommand{\gr}[1]{\textcolor{gray}{#1}}

\resizebox{\textwidth}{!}{
\setlength{\tabcolsep}{3.8pt}

\begin{tabular}{l cccccc cc}
\toprule

\addlinespace[2pt]

\multicolumn{1}{c}{\textbf{Model}}
& \textbf{ArXivMath}
& \textbf{GPQA-D}
& \textbf{SuperGPQA}
& \textbf{ChemBench}
& \textbf{MedXpertQA}
& \textbf{MMLU-Pro}
& \textbf{Avg Perf.}
& \textbf{Avg Cost ($\downarrow$)} \\

\addlinespace[1pt]
\midrule
\addlinespace[1pt]

Qwen3-4B
& 2.26
& 7.89
& 6.13
& 5.31
& 3.46
& 3.97
& 4.84
& 1.33 \\

Qwen3-8B
& 2.90
& 17.58
& 16.29
& 25.86
& 10.95
& 18.31
& 15.31
& 1.92 \\

Qwen3-14B
& 4.50
& \textbf{26.02}
& \textbf{25.22}
& \underline{37.66}
& \underline{13.61}
& \textbf{25.15}
& \underline{22.03}
& 2.54 \\

\addlinespace[1pt]
\midrule
\addlinespace[2pt]

ForkingRL
& 1.63
& 7.58
& 5.46
& 5.39
& 3.67
& 3.39
& 4.52
& 1.43 \\

Search-R1
& 3.89
& 10.08
& 8.85
& 10.55
& 3.63
& 6.35
& 7.22
& 1.04 \\

Query Opening
& 5.01
& 15.78
& 15.53
& 26.56
& 8.63
& 12.60
& 14.02
& 1.07 \\

Prompt Only
& 4.05
& 12.89
& 7.79
& 6.80
& 3.75
& 5.96
& 6.87
& \textbf{0.83} \\

\rowcolor{citeblue!10}
\textbf{\modelname/-4B-SFT}
& 5.04
& 11.88
& 12.52
& 19.53
& 6.74
& 10.08
& 10.96
& \underline{0.89} \\

\rowcolor{citeblue!20}
\textbf{\modelname/-4B}
& \textbf{6.28}
& 19.61
& 17.98
& 32.19
& 10.52
& 14.53
& 16.85
& 0.93 \\


\rowcolor{citeblue!30}
\textbf{\modelname/-8B}
& \underline{5.25}
& \underline{24.30}
& \underline{22.54}
& \textbf{41.41}
& \textbf{16.39}
& \underline{23.26}
& \textbf{22.19}
& 1.84 \\

\midrule
\addlinespace[1pt]

\rowcolor{gray!8}
\dgr{DS-V4-Pro}
& \dgr{6.16}
& \dgr{60.94}
& \dgr{37.92}
& \dgr{63.05}
& \dgr{44.86}
& \dgr{48.93}
& \dgr{43.64}
& \dgr{6.26} \\

\bottomrule
\end{tabular}

}

\end{table}

%% file: texts/apdx_texts/RL_analysis.tex
\section{Analysis on Reinforcement Learning}
\label{apdx:rl_analysis}
\input{figs_tex/fig_rl3}
The main results show that RL substantially improves over the SFT initialization, but this leaves open what behavior is actually acquired during RL.
We therefore examine the evolution of the querying policy while varying the cost penalty
$\lambda \in \{0.05, 0.1, 0.2\}$.
All runs are initialized from \modelname--4B-SFT and otherwise use the same RL setup.

\paragraph{RL learns to compose queries across multiple turns.}
Our SFT data contains only single-query rescue trajectories, so repeated tool use is not directly demonstrated during supervised training.
Nevertheless, \cref{fig:rl_dynamics}a shows that the number of generation turns increases throughout RL for all three values of $\lambda$.
Thus, iterative querying is not specific to the cost coefficient used in our main experiment.
Rather, RL learns to repeatedly alternate between local reasoning and external information acquisition, composing the query action beyond the behavior explicitly provided by SFT.

\paragraph{The cost penalty primarily controls how external computation is allocated.}
While multi-turn behavior emerges across all penalty strengths, $\lambda$ changes the type of calls used within those trajectories.
As shown in \cref{fig:rl_dynamics}b, a smaller penalty ($\lambda=0.05$) gradually shifts probability away from depth-1 queries and permits greater use of more expensive backends.
In contrast, $\lambda=0.1$ and $\lambda=0.2$ maintain a stronger preference for shallow calls.
This suggests that RL and the cost penalty play distinct roles: RL learns how to compose tool interactions, while $\lambda$ calibrates how much computation each interaction should consume.

\paragraph{Excessive cost pressure can suppress useful successful trajectories.}
Increasing $\lambda$ also changes the credit assigned among correct rollouts.
Because rewards are centered within each rollout group, a correct but relatively expensive trajectory can receive negative advantage when its cost-adjusted reward falls below the group mean.
We observe this behavior for $\lambda=0.2$ in \cref{fig:rl_dynamics}c, whereas it is negligible for the smaller penalties.
A very large cost penalty can therefore move beyond encouraging efficient success and begin actively downweighting some successful tool-use trajectories.
We use $\lambda=0.1$ in our main experiments as a middle regime that preserves the emergence of iterative querying while maintaining substantial pressure toward inexpensive external computation.

Overall, these results sharpen the roles of the two training stages.
SFT teaches the model how to issue a query and integrate the returned observation, while RL learns how to organize these actions over an evolving reasoning trajectory and allocate external computation under a cost constraint.

%% file: figs_tex/fig_rl3.tex
\begin{figure}[t]
    \centering

    \begin{subfigure}[b]{0.31\textwidth}
        \includegraphics[width=\linewidth]{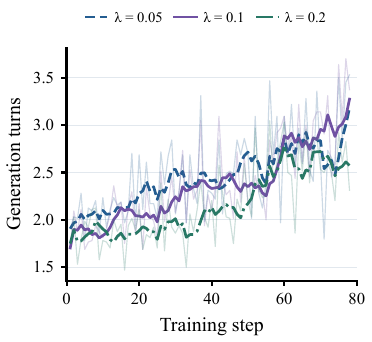}
        \caption{Iterative tool use}
        \label{fig:rl_turns}
    \end{subfigure}\hfill
    \begin{subfigure}[b]{0.31\textwidth}
        \includegraphics[width=\linewidth]{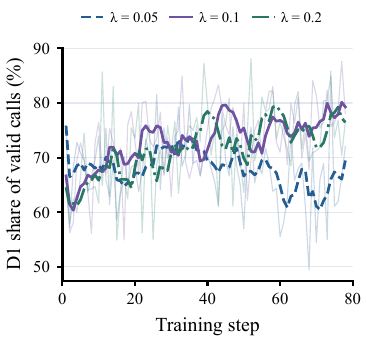}
        \caption{Depth allocation}
        \label{fig:rl_shallow}
    \end{subfigure}\hfill
    \begin{subfigure}[b]{0.31\textwidth}
        \includegraphics[width=\linewidth]{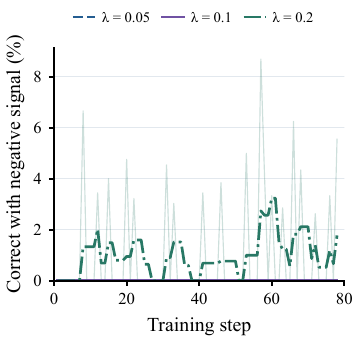}
        \caption{Cost-sensitive credit}
        \label{fig:rl_negadv}
    \end{subfigure}

    \vspace{-0.05in}
    \caption{
    \textbf{RL learns iterative querying across cost penalties, while $\lambda$ controls how aggressively the policy economizes external computation.}
    (a) The number of generation turns increases throughout RL for all tested values of $\lambda$, despite SFT using only single-query rescue trajectories.
    (b) A smaller cost penalty increasingly permits deeper, more expensive calls, whereas larger penalties maintain a stronger preference for depth-1 queries.
    (c) At $\lambda=0.2$, cost-aware reward centering occasionally assigns negative advantage to correct trajectories, indicating that excessive cost pressure can suppress successful but relatively expensive behavior.
    }
    \label{fig:rl_dynamics}
    \vspace{-0.2in}
\end{figure}

%% file: texts/apdx_texts/additional_exp.tex
\section{Additional Experiments}
\label{apdx:ae}
\subsection{Generalization across query backends.}
\label{apdx:gpt}
\input{tables/tool_ablation_table}
To test whether the performance of \modelname--4B is tied to the DeepSeek backend used during training, we replace only the query backend at inference time while keeping the policy and query interface fixed.
For both GPT-5.6 Luna~\citep{openai2026gpt56} and HY3~\citep{tencent2026hy3}, query depths $d\in\{1,2,3\}$ correspond to maximum generation budgets of 128, 512, and 1,536 tokens, respectively.
API costs are computed using the OpenRouter price snapshot collected on September 17, 2026, with input/output prices of \$0.20/\$1.20 per million tokens for GPT-5.6 Luna and \$0.105/\$0.435 for HY3.

As shown in~\cref{tab:backend_transfer}, the fixed policy retains most of its performance after backend substitution, achieving 43.21\% pass@8 with GPT-5.6 Luna and 43.96\% with HY3, compared with 45.96\% using the training-time DeepSeek backend.
Both variants continue to outperform Qwen3-14B at substantially lower serving cost, reducing cost by 56.5\% and 60.6\%, respectively.
These results suggest that the learned policy captures when and how much external assistance to acquire, rather than relying on backend-specific behavior of DeepSeek.

\subsection{Analysis on Leakage}
\input{tables/leakage_table}
\label{apdx:leakage}

A potential concern is that the query tool may improve performance by directly exposing information that makes the answer recoverable without substantial reasoning.
We examine this possibility on 100 randomly sampled hard problems used in our evaluation.
For each problem, we sample 8 rollouts from \modelname/-4B and collect the first successful tool observation whenever a query is issued, yielding 692 observations across 98 problems.
We then evaluate Qwen3-4B with thinking disabled under two conditions: given only the original question, or given the question together with a tool observation.

As shown in \cref{tab:leakage}, adding the observation increases problem-weighted accuracy only from 10.20\% to 12.16\%, a gain of 1.96 percentage points.
These results suggest that the benefit of querying is not primarily explained by direct answer leakage from the tool response.
Instead, the observations are most useful when incorporated into the model's subsequent reasoning process.

\subsection{Additional economic analysis}
\input{figs_tex/fig_state}

Because our cost estimates aggregate API and GPU expenses in USD using prices from OpenRouter and Hyperbolic~\citep{openrouter,hyperbolic}, we examine whether our conclusions are sensitive to changes in either component.
We first consider two representative stress-test scenarios in~\cref{fig:cost_api10,fig:cost_gpu3}.
When API prices increase by $10\times$, API-heavy methods become substantially more expensive, yet \modelname/-4B and \modelname/-8B remain on the Pareto frontier.
Conversely, under a $3\times$ increase in GPU prices, larger local models shift toward higher cost, while our selectively querying models retain favorable accuracy-cost trade-offs.
Together, these scenarios show that the advantage of \modelname/ does not rely on a single pricing regime.

%% file: tables/tool_ablation_table.tex
\begin{table}[ht]
\centering
\caption{
\textbf{Generalization across query backends.}
We transfer the learned querying policy to alternative query backends without retraining.
DeepSeek denotes the training-time backend configuration, which uses different models across query depths, while GPT-5.6 Luna and HY3 replace this backend only at inference time.
We report pass@8 and inference cost over 8 rollouts in milli-dollars (m\$, $10^{-3}$ USD).
Best and second-best results are \textbf{bolded} and \underline{underlined}, respectively.
}
\label{tab:backend_transfer}
\vspace{-0.08in}

\small
\setlength{\aboverulesep}{0pt}
\setlength{\belowrulesep}{0pt}
\renewcommand{\arraystretch}{1.15}

\resizebox{\textwidth}{!}{
\setlength{\tabcolsep}{3.8pt}
\begin{tabular}{ll c ccc c cc c}
\toprule

\addlinespace[2pt]
\multicolumn{1}{c}{\textbf{Model}}
& \multicolumn{1}{c}{\textbf{Query Backend}}
& \textbf{ArXivMath}
& \textbf{GPQA-D}
& \textbf{SuperGPQA}
& \textbf{ChemBench}
& \textbf{MedXpertQA}
& \textbf{MMLU-Pro}
& \textbf{Avg Perf.}
& \textbf{Avg Cost ($\downarrow$)} \\

\addlinespace[1pt]
\midrule
\addlinespace[1pt]

Qwen3-14B & \multicolumn{1}{c}{-}
& 14.46
& 52.63
& \underline{43.85}
& 59.30
& 32.82
& \textbf{46.78}
& 41.64
& 20.35 \\

\addlinespace[1pt]
\midrule
\addlinespace[2pt]

\multirow{3}{*}{\modelname--4B}
& DeepSeek
& 21.78
& \textbf{60.01}
& \textbf{45.65}
& \textbf{70.73}
& \underline{35.37}
& 42.21
& \textbf{45.96}
& \textbf{7.42} \\

& GPT-5.6 Luna
& \textbf{22.39}
& \underline{55.22}
& 43.53
& 64.53
& \textbf{35.59}
& 38.02
& 43.21
& 8.86 \\

& HY3
& \underline{22.19}
& 51.04
& 43.62
& \underline{68.06}
& 34.26
& \underline{44.61}
& \underline{43.96}
& \underline{8.01} \\

\bottomrule
\end{tabular}
}

\end{table}

%% file: tables/leakage_table.tex
\begin{wraptable}{r}{0.43\linewidth}
\vspace{-1.3em}
\centering
\caption{\textbf{Analysis of answer leakage from tool observations.}
We evaluate Qwen3-4B with thinking disabled.}
\label{tab:leakage}
\vspace{-0.4em}

\scriptsize
\setlength{\tabcolsep}{4pt}
\renewcommand{\arraystretch}{1.12}

\resizebox{\linewidth}{!}{%
\begin{tabular}{@{}lcc@{}}
\toprule
\textbf{Input}
& \textbf{Accuracy}
& \textbf{$\Delta$} \\
\midrule
Problem only
& 10.20
& -- \\
Problem + observation
& 12.16
& +1.96 \\
\bottomrule
\end{tabular}
}

\vspace{-0.6em}
\end{wraptable}

%% file: figs_tex/fig_state.tex
\begin{wrapfigure}{r}{0.56\textwidth}
    \centering
    \vspace{-0.2in}
    \captionsetup{font=small,skip=4pt}
    \captionsetup[subfigure]{
    font=small,skip=2pt,justification=centering
    }
    \begin{subfigure}[t]{0.485\linewidth}
        \centering
        \includegraphics[width=\linewidth]{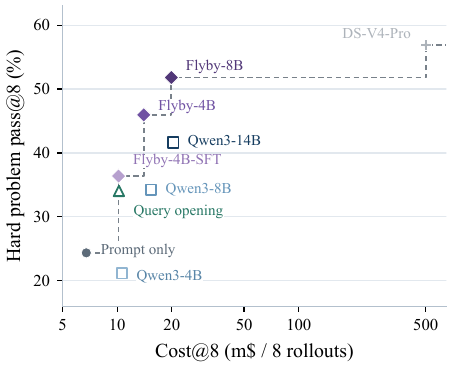}
        \caption{API price $\times10$}
        \label{fig:cost_api10}
    \end{subfigure}\hfill
    \begin{subfigure}[t]{0.485\linewidth}
        \centering
        \includegraphics[width=\linewidth]{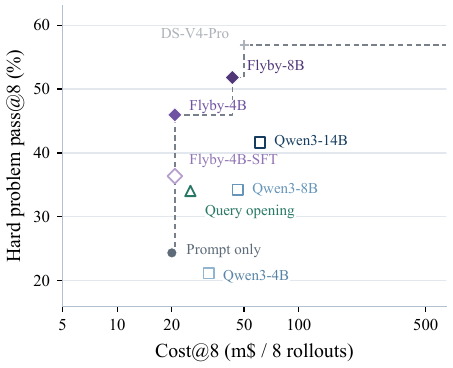}
        \caption{GPU price $\times3$}
        \label{fig:cost_gpu3}
    \end{subfigure}
    \caption{\textbf{Pareto frontiers under representative price perturbations.}
    Cost-performance trade-offs when (a) API prices are increased by $10\times$ and 
    (b) GPU prices are increased by $3\times$.
    }
    \vspace{-0.1in}
    \label{fig:econ}
\end{wrapfigure}

%% file: texts/apdx_texts/examples.tex
\section{Qualitative Examples}
\label{apdx:ex}

\input{figs_tex/fig_qual1}

\cref{fig:qualitative_med} shows how \modelname/-4B uses external information while retaining the reasoning burden itself.
The model first reasons over the clinical presentation and narrows the unresolved issue to the mechanism of hemolysis after mechanical valve replacement.
It then asks a targeted question about the cause of hemolytic anemia in patients with mechanical valves and its relationship to valve function.

The returned reply supplies the missing factual connection: paravalvular leak is a common cause of mechanical hemolysis and is associated with valve dysfunction.
Importantly, the reply neither answers the original multiple-choice question nor mentions transesophageal echocardiography (TEE).
Instead, \modelname/-4B integrates this fact with its preceding reasoning, connects the suspected paravalvular leak to the new murmur and valve dysfunction, and independently infers that TEE is the appropriate next step.
At the state level, the query moves the model out of a knowledge-bottleneck regime, increasing continuation success from 0/8 to 7/8.

\input{figs_tex/fig_qual2}

\cref{fig:qualitative_phy} illustrates the same behavior in a physics problem.
The model initially obtains an implausibly small torque and explicitly recognizes that its calculation may be missing something fundamental.
Rather than requesting a solution to the original problem, it asks for the correct drag formulation for a cylinder in cross-flow and whether the cylinder length enters the expression.

The returned reply provides only the relevant physical relation:
the drag force uses projected frontal area, with $A = dL$ for a cylinder in cross-flow.
This allows the model to identify its own mistake of using the cross-sectional area $\pi r^2$, recompute the drag forces for the three antenna sections, and independently obtain the correct total torque of $7.5425\,\mathrm{N\,m}$.
Correspondingly, continuation success increases from 2/8 before the query to 8/8 afterward.

Together, these examples illustrate the intended behavior of \modelname/-4B.
It reasons locally to identify what information is missing, queries for that specific information, and then resumes the remaining reasoning itself rather than outsourcing the original problem to the external model.

%% file: figs_tex/fig_qual1.tex
\begin{figure}[h]
    \centering
    \includegraphics[width=0.95\linewidth]{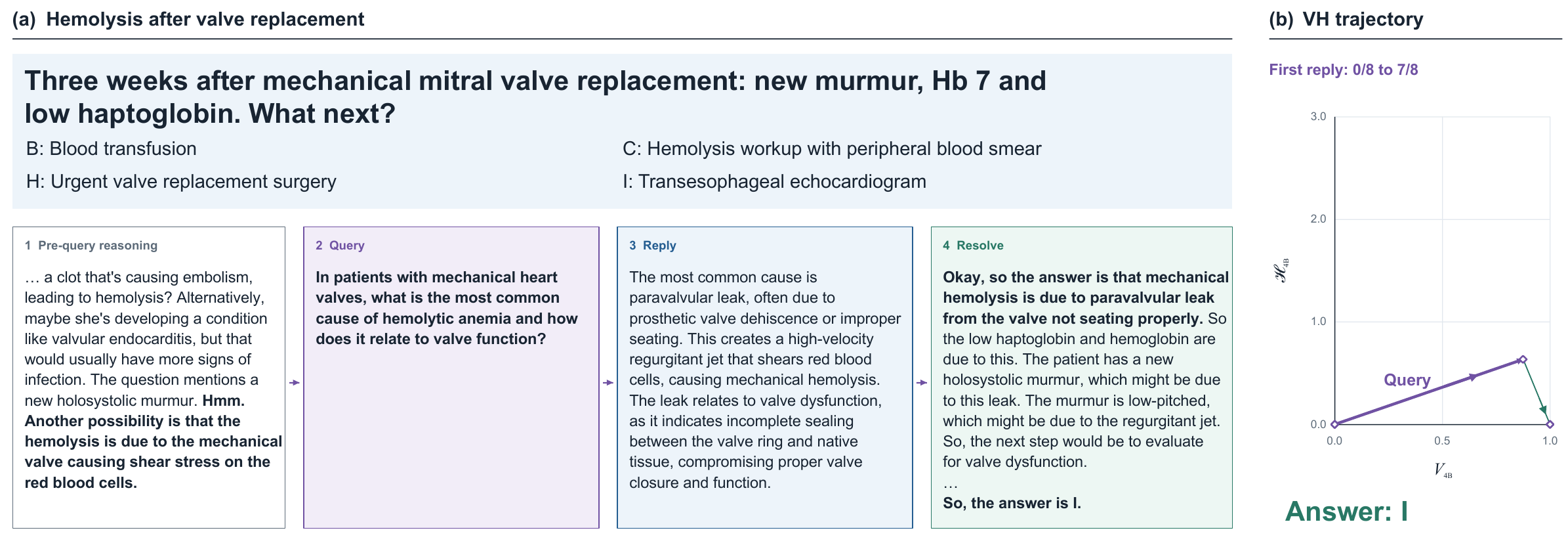}
    \vspace{-0.1in}
    \caption{
    \textbf{Qualitative example of \modelname{}-4B on a medical problem.}
    (a) The original problem and the \modelname{}-4B's reasoning trajectory, consisting of pre-query reasoning, a targeted query, the returned information, and subsequent resolution.
    (b) The corresponding reasoning-state trajectory probed with Qwen3-4B.
    The query supplies the missing factual connection between mechanical hemolysis and paravalvular leak, increasing continuation success from 0/8 to 7/8.
    }
    \label{fig:qualitative_med}
\end{figure}

%% file: figs_tex/fig_qual2.tex
\begin{figure}[ht]
    \centering
    \includegraphics[width=0.95\linewidth]{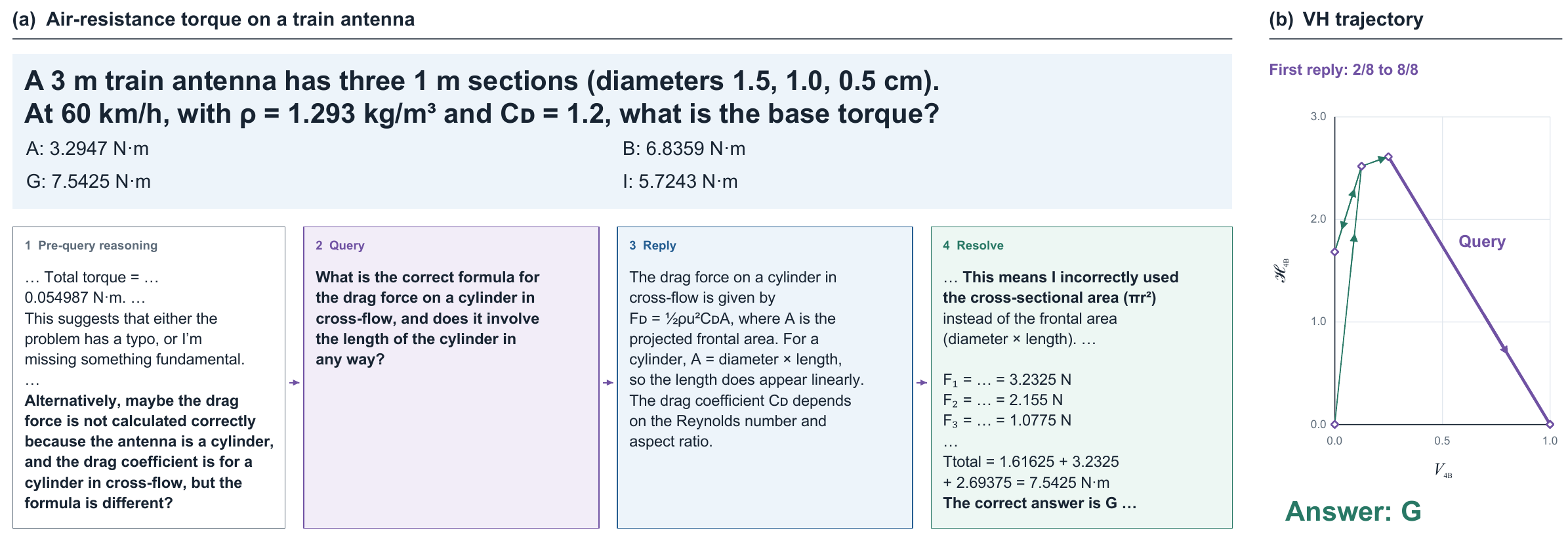}
    \vspace{-0.1in}
    \caption{
    \textbf{Qualitative example of \modelname{}-4B on a physics problem.}
    (a) The original problem and the \modelname{}-4B's reasoning trajectory.
    After obtaining an implausibly small torque, the model queries the correct drag formulation for a cylinder in cross-flow.
    The returned information identifies the projected frontal area as diameter $\times$ length, after which the model corrects its calculation and reaches the answer independently.
    (b) The corresponding reasoning-state trajectory probed with Qwen3-4B.
    Continuation success increases from 2/8 to 8/8 after the query.    }
    \label{fig:qualitative_phy}
\end{figure}

%% file: texts/apdx_texts/limit.tex
\section{Limitations and future directions}
\label{apdx:lim}
In this work, \modelname/ relies on stronger external models to overcome knowledge bottlenecks. This introduces two limitations. First, external API access may not always be available or desirable due to privacy, latency, and deployment constraints. Second, external models can themselves produce incorrect or hallucinated information, which may propagate into subsequent reasoning.

An important direction for future work is privacy-aware selective querying. By learning to selectively reveal or fragment context, local models could acquire external information while minimizing the exposure of sensitive information. Training querying policies to balance information utility and privacy could enable local models to benefit from frontier intelligence without disclosing the full problem context.